\documentclass{article}

\usepackage{PRIMEarxiv}

\usepackage[utf8]{inputenc} % allow utf-8 input
\usepackage[T1]{fontenc}    % use 8-bit T1 fonts
\usepackage{hyperref}       % hyperlinks
\usepackage{url}            % simple URL typesetting
\usepackage{booktabs}       % professional-quality tables
\usepackage{amsfonts}       % blackboard math symbols
\usepackage{nicefrac}       % compact symbols for 1/2, etc.
\usepackage{microtype}      % microtypography
\usepackage{lipsum}
\usepackage{fancyhdr}       % header
\usepackage{graphicx}       % graphics
\graphicspath{{media/}}     % organize your images and other figures under media/ folder

\usepackage{listings}
\usepackage{latexsym}
\usepackage{amsfonts}
\usepackage{amsmath}
\usepackage{graphicx}
\usepackage{qtree}
\usepackage{pict2e}
\usepackage{eepic}
\usepackage{color}
\usepackage{moreverb}
\usepackage{fancyvrb}
\usepackage{tabularx}
\usepackage{colortbl}
\usepackage{acronym}
\usepackage{multirow}
\usepackage{longtable}
\usepackage{paralist}
\usepackage{url}
\usepackage{xspace}
\usepackage{multirow}			% multirow nas tabelas
\usepackage{subcaption}
\usepackage{algorithm}
\usepackage{algorithmicx}
\usepackage{algpseudocode}
\usepackage{verbatim}			% verbatim, comment, etc.
\usepackage{array}				% Extens„o de tabular e array
\usepackage{colortbl}			% Tabelas com cores
\usepackage{tikz-qtree}
\usepackage{rotating}
\usepackage{etoolbox}
\usepackage{indentfirst}
\usepackage{booktabs}
\usepackage{epstopdf}
\usepackage{bm}
\usepackage{natbib}

\acrodef{AI}{Artificial Intelligence}
\acrodef{API}{Application Programming Interface}
\acrodef{AutoML}{Automatic Machine Learning}
\acrodef{CA}{Conversational Agent}
\acrodef{CNN}{Convolutional Neural Network}
\acrodef{DOM}{Document Object Model}
\acrodef{GPU}{Graphics Processing Unit}
\acrodef{HTML}{HyperText Markup Language}
\acrodef{HTTP}{Hypertext Transfer Protocol}
\acrodef{IMDb}{International Movie Database}
\acrodef{IMSDb}{Internet Movie Script Database}
\acrodef{IP}{Internet Protocol}
\acrodef{IST}{Instituto Superior Técnico}
\acrodef{JSON}{JavaScript Object Notation}
\acrodef{LSTM}{Long-Short Term Memory}
\acrodef{LTR}{Learn To Rank}
\acrodef{NER}{Named Entity Recognition}
\acrodef{NLP}{Natural Language Processing}
\acrodef{NLTK}{Natural Language Toolkit}
\acrodef{NL}{Natural Language}
\acrodef{NMT}{Neural Machine Translation}
\acrodef{OCR}{Optical Character Recognition}
\acrodef{OMDb}{Open Movie Database}
\acrodef{OOD}{Out Of Domain}
\acrodef{POS}{Part-Of-Speech}
\acrodef{QA}{Question and Answer}
\acrodef{RNN}{Recurrent Neural Networks}
\acrodef{SAX}{Simple API for XML}
\acrodef{seq2seq}{Sequence-to-Sequence}
\acrodef{SSD}{Say Something Deep}
\acrodef{SSS}{Say Something Smart}
\acrodef{StAX}{Streaming API for XML}
\acrodef{TF-IDF}{Term Frequency-Inverse Document Frequency}
\acrodef{XML}{eXtensible Markup Language}
\acrodef{YAML}{YAML Ain't Markup Language}

\definecolor{notesColor}{RGB}{134, 179, 0}

\graphicspath{{./figs/}}

\newcommand{\sys}{GEN\xspace}
\newcommand{\du}{D\&A\xspace}

\newcommand{\McQGE}{\textsc{Monserrate}\xspace}

\acrodef{IR}{Information Retrieval}
\acrodef{IE}{Information Extraction}
\acrodef{WWW}{World Wide Web}
\acrodef{QA}{Question Answering}
\acrodef{QG}{Question Generation}
\acrodef{NP}{Noun Phrase}
\acrodef{VP}{Verb Phrase}
\acrodef{LAT}{Lexical Answer Type}
\acrodef{IMDB}{Internet Movie Database}
\acrodef{MRR}{Mean Reciprocal Rank}
\acrodef{NE}{Named Entity}
\acrodefplural{NE}[NEs]{Named Entities}
\acrodef{TREC}{Text REtrieval Conference}
\acrodef{SVM}[SVM]{Support Vector Machine}
\acrodef{SDS}[SDS]{Spoken Dialog System}
\acrodef{LFPR}[LFPR]{Logical Framework for Probabilistic Reasoning}
\acrodef{ASR}[ASR]{Automatic Speech Recognition}
\acrodef{TTS}{Text-To-Speech}
\acrodef{BDI}[BDI]{Believes Desires Intention}
\acrodef{SLU}[SLU]{Spoken Language Understanding}
\acrodef{HMM}[HMM]{Hidden Markov Model}
\acrodef{LUP}[LUP]{Language Understanding Platform}
\acrodef{NLU}[NLU]{Natural Language Understanding}
\acrodef{XML}[XML]{Extensible Markup Language}
\acrodef{SQL}[SQL]{Structured Query Language}
\acrodef{ITS}[ITS]{Intelligent Tutoring System}
\acrodef{ECA}{Embodied Conversational Agent}
\acrodef{NER}{Named Entity Recognition}
\acrodef{NLP}{Natural Language Processing}
\acrodef{VTA}{Virtual Teaching Assistant}
\acrodef{POS}{Part of Speech}
\acrodef{SRL}{Semantic Role Labeler}
\acrodef{MOOC}{Massive Open Online Course}
\acrodef{DB}{Database}
\acrodef{MSRP}{Microsoft Research Paraphrase Corpus}
\acrodef{QGSTEC}{Question Generation Shared Task \& Evaluation Challenge}
\acrodef{QAD}{Question Answer Dataset}
\acrodef{AMT}{Amazon Mechanical Turk}
\acrodef{NLG}{Natural Language Generation}
\acrodef{EACS}{Embedding Average Cosine Similarity}
\acrodef{GMS}{Greedy Matching Score}
\acrodef{STCS}{SkipThought Cosine Similarity}
\acrodef{VECS}{Vector Extrema Cosine Similarity}
\acrodef{EWAF}{Exponentially Weighed Average Forecast}
\acrodef{WMA}{Weighed Majority Algorithm}

\title{Using Implicit Feedback to Improve Question Generation}
\author{
Hugo Rodrigues\\
Instituto Superior T\'{e}cnico, Universidade de Lisboa\\INESC-ID\\Language Technologies Institute, Carnegie Mellon University\\
\texttt{hugo.p.rodrigues@tecnico.ulisboa.pt}
\And
Eric Nyberg\\
Language Technologies Institute, Carnegie Mellon University\\
\texttt{ehn@cs.cmu.edu}
\And
Luisa Coheur\\
Instituto Superior T\'{e}cnico, Universidade de Lisboa\\INESC-ID\\
\texttt{luisa.coheur@tecnico.ulisboa.pt}
}
\begin{document}
\maketitle
\begin{abstract}
\ac{QG} is a task of \ac{NLP} that aims at automatically generating questions from text. Many applications can benefit from automatically generated questions, but often it is necessary to curate those questions, either by selecting or editing them. This task is informative on its own, but it is typically done post-generation, and, thus, the effort is wasted. In addition, most existing systems cannot incorporate this feedback back into them easily. In this work, we present a system, \sys, that learns from such (implicit) feedback. Following a pattern-based approach, it takes as input a small set of sentence/question pairs and creates patterns which are then applied to new unseen sentences. Each generated question, after being corrected by the user, is used as a new seed in the next iteration, so more patterns are created each time. We also take advantage of the corrections made by the user to score the patterns and therefore rank the generated questions. Results show that \sys is able to improve by learning from both levels of implicit feedback when compared to the version with no learning, considering the top 5, 10, and 20 questions. Improvements go up from 10\%, depending on the metric and strategy used.
\end{abstract}

%!TEX root = main.tex

\section{Introduction}\label{sec:monserrate}

\ac{QG} is the Natural Language Processing task of automatically generating questions from unseen sentences/text. Manually creating questions is a time-consuming task, and many different scenarios could benefit from automatically created questions. Educational settings, in which teachers have to create questions to assess their students, and the creation of a knowledge base for a chatbot that performs, for instance, customer support, are a few examples of how \ac{QG} can be applied. In a real \ac{QG} setting, users are likely required to correct at least some of the obtained questions. Typically, these corrections are wasted, although being informative. In other words, they can be seen as implicit feedback on the system's performance. In this work we explore the concept of taking advantage of this implicit feedback to improve the performance of our proposed \ac{QG} system. 

To make use of such data, the system needs to be capable of incorporating it online. Neural networks have recently drawn some attention in the \ac{QG} field \citep{Serban16,Du17,Wang17}, but they can hardly benefit from such small learning increments. The more traditional approaches to \ac{QG} rely in rules or patterns, which are used to transform input sentences into new questions \citep{Kalady10, Heilman10,Ali10, Varga10}. However, most of them utilize hand-crafted patterns, which not only requires linguists to create them, but also do not give room to adapting the systems on the long term. In other fields, like relation extraction, some works studied automatic acquisition of patterns \citep{Ravichandran02, Velardi13}, a strategy employed, to the best of our knowledge, by a single work in \ac{QG}: The Mentor \citep{Curto11}. Inspired by it, we designed a pattern-based \ac{QG} system that automatically learns patterns from question/sentence seeds, which are used to generate questions from new sentences. We expand the original idea by employing multiple matching strategies and semantic features, while including the concept of using implicit feedback as well. 

In this paper, we present \sys\footnote{This paper deeply expands our previous work \citep{Rodrigues18}, where we showed a preliminary architecture and results of \sys.}, our \ac{QG} system, which is able to take advantage of the implicit feedback given by the expert when editing the generated questions, incorporating it back on the system in two ways. First, the edited questions, along with their source sentence, are used as new seeds, thus enlarging the pool of available patterns. Secondly, the corrections made by the expert are used as a tool to score the corresponding patterns, which are used indirectly to rank the generated questions. Contrasting with deep learning methods that typically require larger datasets for training, \sys only needs a small set of question/sentence seeds to be bootstraped. In addition, short loops of human feedback can improve its performance, unlike state of the art counterparts.

The paper is organized as follows: in Section~\ref{sec:evalRW} we present related work, and in Section~\ref{sec:sys} we describe our system, \sys. In Section~\ref{sec:es} we detail the experimental setup and in Sections~\ref{sec:results} through~\ref{sec:weightSquad} we report our experiments and results. Finally, we highlight the main conclusions and suggest future work in Section~\ref{sec:conc}.

%%%%%%%%%%%%%%%%%%%%%%%%%
%%%%%%%%%%%%%%%%%%%%%%%%%
%%%%%%%%%%%%%%%%%%%%%%%%%
%%%%%%%%%%%%%%%%%%%%%%%%%
%%%%%%%%%%%%%%%%%%%%%%%%%

\section{Related Work} \label{sec:evalRW}

%%%%

In what concerns \ac{QG} systems, most rely on hand-crafted rules \citep{Ali10, Varga10, Pal10} or tree operations on parse trees to transform sentences into questions \citep{Kalady10, Heilman10, Heilman09, Wyse09}. Examples of the former approach include the systems introduced by \citet{Mazidi15}, \citet{Mannem10}, \citet{Ali10}, or \citet{Varga10} which take advantage of hand-crafted patterns. For example, a designed rule like \texttt{NP$_1$ VB NP$_2$ $\rightarrow$ Wh-word VB NP$_2$?} transforms a simple subject-verb-object sentence into a question about the subject of that sentence. Most of these systems rely on constituent parsers, but some take a step further. For instance, \citet{Mazidi15} use lexico-syntactic patterns based in dependency parsers, while \citet{Mannem10} take advantage of a \ac{SRL}. Approaches based on tree operations are illustrated by the work described by~\citet{Heilman10}, directly operating on the parse trees to transform the sentences into questions. This system also adopts a strategy of over-generation, ranking the generated questions afterwards. Also following in the hand-crafted paradigm, the system presented in \cite{Labutov18} asks the crowd to create questions that are likely to generalize, regarding entities of type Person and Locations. 

A totally different approach was followed by THE-MENTOR \citep{THE-MENTOR2,Curto12} and \sys \citep{Rodrigues18}, which automatically learn patterns from a set of sentence/question/answer triples (the seeds). In both systems, the pattern is created from the sentences that contain both the question and answer terms, and the shared tokens among the sentence and the question tell how one can be transformed into the other, generating questions of the original type. While THE-MENTOR only uses syntactic cues, \sys allows different types of pattern matching, from syntactic to semantic. This type of approach is based in similar strategies used in other domains, like QA \citep{Ravichandran02}.

More recently, with the technological advancement of GPUs, and the creation of large datasets (like SQuAD \citep{Rajpurkar16}) to help them, neural networks gained popularity in different domains, putting state of the art results. When it comes to \ac{QG}, it became possible to successfully train large neural networks with the goal of generating questions \citep{Serban16,Indurthi17,Du17,Wang17,Zhou18,Subramanian18,sun-etal-2018-answer,MixQG}. These works apply a sequence-to-sequence approach \citep{Sutskever14}, with most of them requiring the answer span as an input of the network \citep{Yuan17,Zhou18,Liu19, Subramanian18}. Other works use more structured inputs, like ontology triples that describe factual relationships between two entities \citep{Serban16, Indurthi17,QGTriplets}. However, this imposes a limitation on the applicability of the approach, given that such data (ontology triples) is not widely available. Some works take a single sentence or paragraph alone as input. The work of \citet{Du17} show how to generate questions from unstructured text without requiring the answer span as an input. This approach is more realistic and mimics better what previous \ac{QG} systems did in the past, as the answer span is typically not available. The authors exploit a sentence- and a paragraph-level model, where the latter encodes context information to better inform the generation process. \citet{Kumar18} also use a similar approach, with the difference being that the loss function is adapted to integrate typical evaluation metrics like ROUGE and BLEU. The additional input of the answer is also behind dual strategies, that train a model aimed at solving two tasks jointly. In this case, \ac{QA} is the joint problem chosen, given the similarity between the task and \ac{QG}. Despite their distinct research areas, \ac{QG} and \ac{QA} are seen as complementary aspects of the same problem. In fact, \ac{QA} has a long tradition in \ac{NLP} research on its own, contributing with widely know systems as Wolfram Alpha\footnote{\url{https://www.wolframalpha.com}}, IBM Watson \citep{watson}, and others \citep{Soubbotin01,Mendes13,Fader14,Baudi15}, including more recent neural network models, like BERT \citep{Devlin18}\footnote{There is also significant research on non-factoid \ac{QA} \citep{Bilotti07,Higashinaka08,Surdeanu11}, but it is out of the scope of this paper to explore that topic in detail.}. Thus, for instance, \citet{Wang17} use an encoder-decoder model that is conditioned on a secondary input, either the answer (\ac{QG}), or the question (\ac{QA}), which is used to solve both tasks, iteratively. Another example, by \citet{Tang17}, looks at \ac{QA} as a ranking problem, i.e., it uses the answer selection task only, while the \ac{QG} part of the model does use the answer as an input, while trying to minimize a common loss function for both tasks. 

Attention mechanisms, widely used in typical sequence-to-sequence models, have been used to create Transformers \citep{Vaswani17}. This is also a sequence-to-sequence model that discards the recurrent units of the previous models (LSTM \citep{Hochreiter97} or GRU \citep{Cho14}), implementing a stack of attention mechanisms instead, overcoming some of their limitations, both conceptual (long input sequences) and technical (training resources needed). Transformers established a new standard on neural networks state of the art, being the base for BERT \citep{Devlin18}, a model that can be successfully fine-tuned for a variety of tasks, and that was recently used in \ac{QG} \cite{chan-fan-2019-bert}. Currently, we can find neural \ac{QG} being applied to a panoply of scenarios, such as education settings \citep{QGEducation}, conversations \citep{QGConversations} or specific languages \citep{QGChinese}. Two recent surveys are also available, one focusing in \ac{QG} in general terms \citep{surveyGeneral}, the other targeting neural \ac{QG} \citep{surveyDeepQG}.

%%%%%%%%%%%%%%%%%%%%%%%%%%%%%%
%%%%%%%%%%%%%%%%%%%%%%%%%%%%%%
%%%%%%%%%%%%%%%%%%%%%%%%%%%%%%
%%%%%%%%%%%%%%%%%%%%%%%%%%%%%%
%%%%%%%%%%%%%%%%%%%%%%%%%%%%%%

\section{\sys}\label{sec:sys}
\begin{figure}[htbp]
\begin{center}
\includegraphics[width=\textwidth]{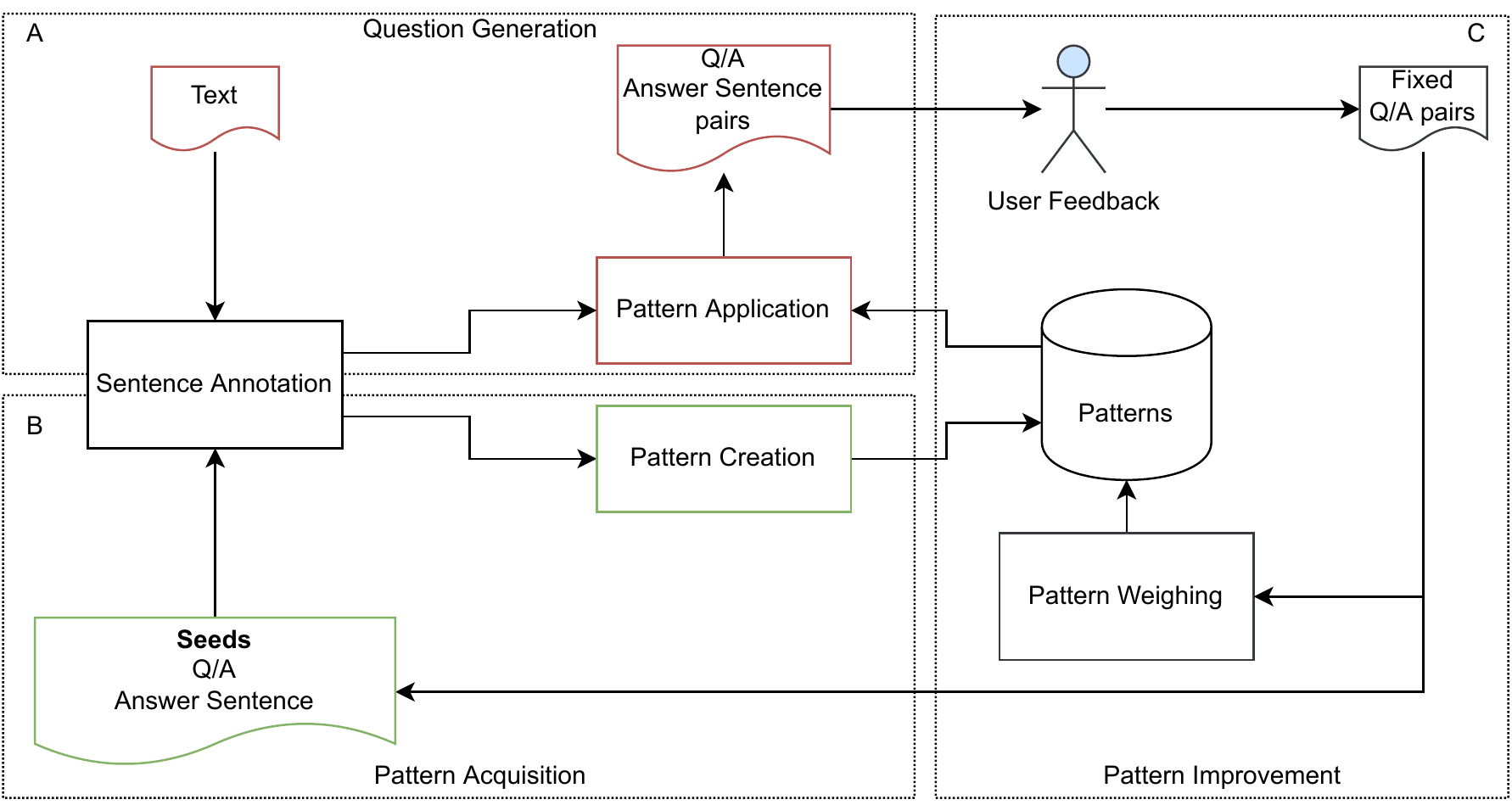}
\caption{Proposed solution for \sys, depicting all steps of the process: Pattern Acquisition (\texttt{B}), Question Generation (\texttt{A}), and Pattern Improvement (\texttt{C}).}
\label{fig:SistArch}
\end{center}
\end{figure}

In this section we will describe \sys, our pattern-based \ac{QG} system that automatically learns patterns from question/sentence seeds. Then, these patterns are used to generate questions from new sentences. Figure~\ref{fig:SistArch} depicts the overall idea. In a first step, semantic patterns are created based on the seeds, in the Pattern Acquisition step (\texttt{B}, in the bottom left of Figure~\ref{fig:SistArch}). This step is detailed in Section~\ref{sec:patternAcq}. The top left part of the figure (\texttt{A}) respects to the application of those patterns to new unseen sentences, resulting in new generated questions (Section~\ref{sec:patternApp}). In Figure~\ref{fig:SistArch}, on the right (\texttt{C}), is also depicted the idea of using the implicit feedback from experts to create new seeds and weigh patterns across multiple iterations. This happens along the generation process, and is discussed in Section~\ref{sec:feedback}.

\subsection{Pattern Acquisition}
\label{sec:patternAcq}

We define a pattern $\mathcal{P}(Q, A, S)$ as a tuple $<Q/A, \mathcal{PA}(S), \widehat{align(Q/A, S)}>$, where:
\begin{itemize}
\item $Q$, $A$ and $S$ belong to a seed and correspond to the Q/A pair and the answer sentence, respectively; 
\item $\mathcal{PA}(S)$ is a predicate-argument structure that captures the different information about the answer sentence $S$; 
%each predicate in $S$ will lead to a predicate-argument structure, and, thus, to a pattern;
\item $\widehat{align(Q/A, S)}$ is the alignment between the tokens in $Q/A$ and $S$.
\end{itemize}

This tuple contains all the needed information to create a new question $Q'$ from a new unseen sentence $S'$, as we will see in next section. In the following subsections we detail how these patterns are generated.

\subsubsection{Sentence Annotation and Predicate-Arguments}
As for sentence annotation, each sentence is enriched with information from multiple sources. At the token level, each token $t$ is annotated with the following:
\begin{itemize}
\item Named Entities: if a word or multi-word expression is detected as a \ac{NE} (we used regular expressions to extract dates in addition to the Stanford Named Entity Recognizer \citep{Finkel05}), it will be collapsed into a single token, and will be tagged with the \ac{NE} type (for instance, \texttt{Person} or \texttt{Location});
\item WordNet: \sys identifies the synsets to which each token belongs, given by WordNet \citep{Miller95};
\item Verb Sense: if the token is a verb, its frame and/or class is noted, according to FrameNet \citep{Baker03} and VerbNet \citep{Kipper00}, respectively;
\item \ac{POS}: the token is labeled with its \ac{POS} tag (parsed by the Stanford parser);
\item Word Embedding: the token is associated with its word embedding vector (by using Word2Vec \citep{Milokov13}).
\end{itemize}

Then, at the sentence level, we use Stanford constituent and dependency parsers \citep{Klein03,Marneffe06} to create both constituent and dependency trees, and Senna \ac{SRL} \citep{Collobert11} to obtain the semantic roles. The \ac{SRL} provides us predicate-argument structures that we use to capture the semantic information of the answer sentence. That is, each predicate identified by the \ac{SRL} in the answer sentence will generate a triple (the ``predicate-argument structure'') composed of:
\begin{itemize}
\item the root of the predicate (a verb);
\item a set of arguments (associated to that verb);
\item a set of subtrees, extracted from both the constituent and dependency trees, so that each subtree captures the arguments of the predicate. 
\end{itemize}

\begin{figure}[h!t]
\centering
\begin{subfigure}[b]{0.50\textwidth}
\includegraphics[width=\textwidth]{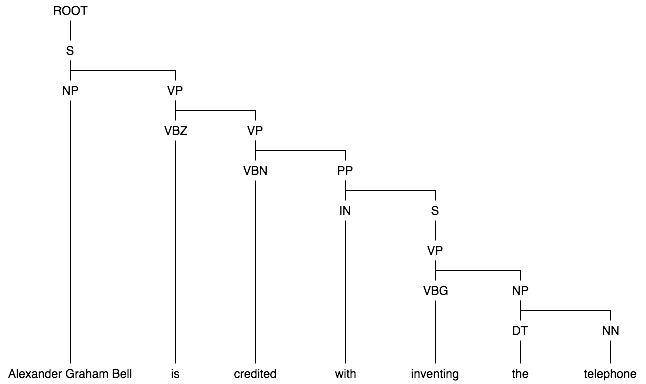}
\caption{}
\label{fig:constTree}
\end{subfigure}
\begin{subfigure}[b]{0.45\textwidth}
\includegraphics[width=\textwidth]{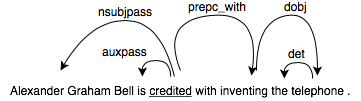}
\caption{}
\label{fig:depTree}
\includegraphics[width=0.9\textwidth]{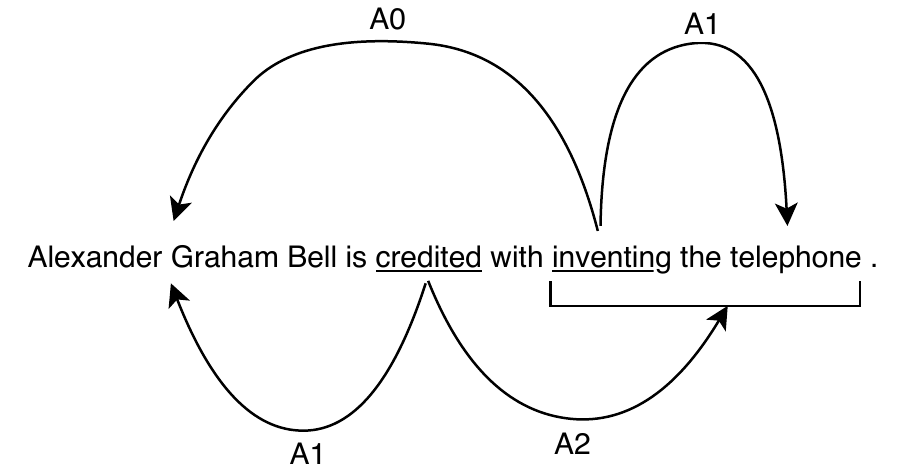}
\caption{}
\label{fig:srlTree}
\end{subfigure}
\caption{Parse trees for the answer sentence \textit{Alexander Graham Bell is credited with inventing the telephone} obtained with (a) Stanford syntactic parser, (b) Stanford dependency parser, (c) Senna Semantic Role Labeler.}
\label{fig:trees}
\end{figure}

\begin{figure}[tbp]
\begin{center}
\includegraphics[width=\textwidth]{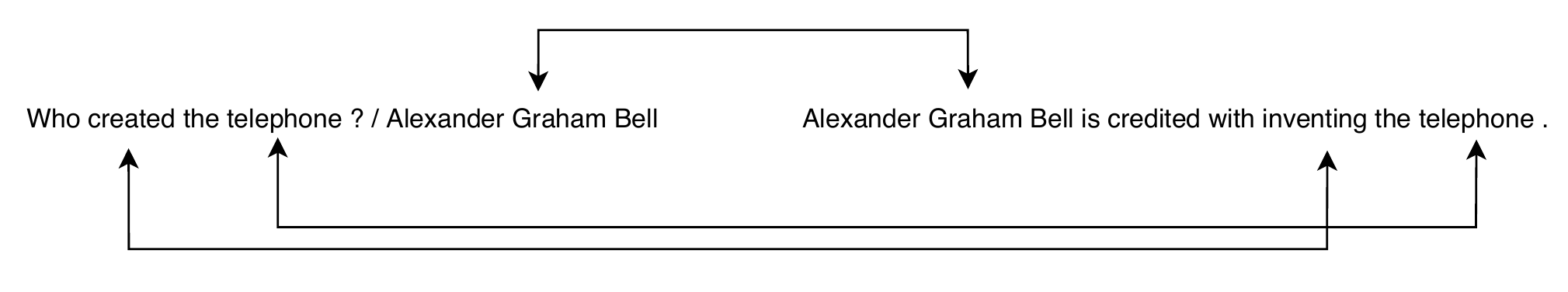}
\caption{An alignment between the Q/A pair (\textit{Who created the telephone?/Alexander Graham Bell}) and the answer sentence \textit{Alexander Graham Bell is credited with inventing the telephone.}}
\label{fig:rules}
\end{center}
\end{figure}

Consider Figure~\ref{fig:trees} and the answer sentence $S$, \textit{Alexander Graham Bell is credited with inventing the telephone}. According to the \ac{SRL}, the predicate \textit{credited} will lead to the predicate-argument:
\[<credited, \{A_1, A_2\}, ST(A_1) \cup ST(A_2)>,\]
where $ST(A_1)$ contains all the subtrees that capture argument \texttt{A1} (\textit{Alexander Graham Bell}) and $ST(A_2)$ contains the subtrees that capture argument \texttt{A2} (\textit{with inventing the telephone}). For instance, examples of the latter are the following subtrees:
\begin{itemize}
\item \texttt{(PP (IN with) (S (VP (VBG inventing)  (NP (DT the) (NN telephone)))))},
\item \texttt{(S (VP (VBG inventing)  (NP (DT the) (NN telephone))))},
\item \texttt{inventing $\xrightarrow{\text{\fontsize{3pt}{5pt}\selectfont dobj}}$ telephone, $\xrightarrow{\text{\fontsize{3pt}{5pt}\selectfont det}}$ the}. 
\end{itemize}

\subsubsection{Question-Answer Alignment}
In what concerns the alignment between the Q/A pair and the answer sentence, Figure~\ref{fig:rules} depicts this idea, by showing an alignment between the tokens of the Q/A pair with the ones of the answer sentence. Notice that the Wh-word is not aligned with the answer sentence. As we will see, this is why we need the Q/A pair in the pattern, so that the Q/A structure guides the generation of the new question. To perform such alignment, we require that:
\begin{enumerate} 
\item [$R_1$:] the content of one sentence is included into the other. Without loss of generality, consider that the content of the $Q/A$ seed is contained in the answer sentence $S$;
\item [$R_2$:] all tokens in $Q/A$ are aligned with one and only one token in $S$;
\item [$R_3$:] no token in $S$ is associated with more than one token in $Q/A$.
\end{enumerate}

We find the best alignment between the $Q/A$ pair and the answer sentence $S$, $\widehat{align(Q/A, S)}$, among all possible alignments, by satisfying Equation \ref{eq:align1}.

\begin{equation}
\widehat{align(Q/A, S)} = \operatorname*{arg\,max}_{a \in \mathcal{A}}~score(a),
\label{eq:align1}
\end{equation}
where $\mathcal{A}$ is the set of all possible alignments between $Q/A$ and $S$ that respect the previous requirements ($R_1-R_3$), and $score(a)$ is the score given to alignment $a$. This score is given by the sum of the scores of each individual alignment, token-wise:

\begin{equation}
score(a) = \sum_{t^i \in T_{Q/A}, t^j \in T_S} score(align(t^i, t^j)),
\label{eq:align2}
\end{equation}
where $align(t^i, t^j)$ is an alignment between tokens $t^i$ and $t^j$, and $T_{Q/A}$ and  $T_{S}$ are the set of tokens in $Q/A$ and $S$, respectively.

The alignment $\widehat{align(Q/A, S)}$ maximizes the alignment between the tokens from $Q/A$ and $S$. As virtually any token in $Q/A$ could align with a token in $S$, choosing the best set of token alignments is similar to the assignment problem, in which one is aiming at optimizing an utility function over a set of assignments. So, being $M$ a matrix of dimension $|t_{Q/A}| \times |t_S|$, each position $M_{ij}$ contains the alignment score for $align(t^i, t^j)$.

% -- usually one tries to minimize the cost of doing $X$ jobs by using $X$ workers, each with a known hourly rate for each job. In our case we intend to pick the best pairwise token alignments that maximize the overall quality of the alignment $\widehat{align(Q/A, S)}$, given that we only use each token once. Therefore, instead of jobs and workers, we have tokens belonging to $Q/A$ and $S$, each with a $score(align(t^i, t^j))$. So, being $M$ a matrix of dimension $|t_{Q/A}| \times |t_S|$, each position $M_{ij}$ contains the alignment score for $align(t^i, t^j)$.

The Hungarian algorithm \citep{Kuhn55} is a combinatorial optimization algorithm designed to solve the assignment problem. We adapted the Hungarian algorithm to our problem. %The assignment problem usually takes an equal number of jobs and workers, but an adaptation is possible by adding the necessary dummy lines/columns if the matrix is not square.  
The original problem tries to minimize the utility function, while we are trying to maximize the value of the overall alignment. To make for this adaptation, we convert the values to have a minimization problem instead, replacing each cell by $max - M_{ij}$, where $max$ is the maximum value present in the whole matrix.

Notice that for both $Q/A$ and $S$ we only consider for alignment tokens that are not stopwords, as the exact stopwords are unlikely to appear in the a new given sentence and are, thus, irrelevant to establish a relationship between them.

\subsubsection{Token Matching}
The missing piece for the alignment process is how tokens themselves are aligned and scored. We designed a set of functions to this end that compare two tokens and give a score on their similarity based on the different tokens' features, from the annotation process previously mentioned. Generically, these functions are referred to as function $equiv$, and its outcome is used to score token alignments in $align(t^i, t^j)$. They are defined as follows:

\paragraph{\textbf{Lexical}} The first one, $equiv_L$, performs a lexical comparison of the two tokens:
\begin{align}
equiv_{L}(t^i, t^j) =
\begin{cases}
1\hphantom{.75} & \text{if } t^i = t^j\\
0.75  & \text{if } t^i \not= t^j \wedge lemma(t^i) =  lemma(t^j)\\
0 & \text{otherwise}
\end{cases}
\label{eq:lexical}
\end{align}

\paragraph{\textbf{Verb sense}}
$equiv_{VB}$ compares the sense of two tokens if they are both verbs and are related according to SemLink\footnote{\url{http://verbs.colorado.edu/semlink}}. This resource is a mapping between PropBank \citep{Palmer05}, VerbNet and FrameNet. If the two tokens belong to the same set in any of the resources, they are considered to match.
\begin{align}
equiv_{VB}(t^i, t^j) = 
\begin{cases}
0.75\hphantom{.0} & \text{if } sense(t^i) = sense(t^j)\\
0 & \text{otherwise}
\end{cases}
\label{eq:verb}
\end{align}

For example, \textit{make} and \textit{build} both belong to the frame \texttt{Building} of FrameNet, which would make the function return $0.75$ for those tokens. 

\paragraph{\textbf{\acl{NE}}}
$equiv_{NE}$ compares the tokens' content, if both tokens are \acp{NE} of the same type\footnote{At generation time the function is replaced by $type(t^i) = type(t^j)$, as are not trying to align the same entity, but rather to find an entity of the same type.}.
\begin{align}
equiv_{NE}(t^i, t^j) = 
\begin{cases}
1\hphantom{.0} & \text{if } t^i = t^j\\ 
0.9 & includes(t^i, t^j)\\
0 & \text{otherwise}
\end{cases}
\label{eq:NE}
\end{align}

The function $includes(t^i, t^j)$ uses a set of rules (based on regular expressions) to determine if two tokens are referring to the same entity, or if two tokens represent the same date. For instance, both \textit{Obama} and \texttt{D01 M01 Y2014} are included in \textit{Barack Obama} and \texttt{M01 Y2014}, respectively.

\paragraph{\textbf{WordNet}}
$equiv_{WN}$ matches two tokens if their path distance traversing WordNet synsets is below a manually defined threshold. We compute the path distance by traversing the synsets upwards until finding the least common subsumer \citep{Resnik95}. For each node up, a decrement of 0.1 is awarded, starting at 1.0.
\begin{align}
equiv_{WN}(t^i, t^j) = 
\begin{cases}
1 \hphantom{.0} & \text{if } syn(t^i) = syn(t^j)\\
x & \text{if } syn\big(hyp(t^i)\big) \supset hyp(t^j)\\
x & \text{if } hyp(t^i) \subset syn\big(hyp(t^j)\big)\\
0 & \text{otherwise},
\end{cases}
\label{eq:WN}
\end{align}
where $syn(t)$ gives the synset of the token $t$, $hyp(t)$ gives the hypernyms of $t$, and $x = 1 - max(n\times0.1, m\times0.1)$, with $n$ and $m$ being the number of nodes traversed in the synsets of $t^i$ and $t^j$ respectively. If no concrete common subsumer is found, then 0 is the result returned. For example, \textit{feline} and \textit{cat} have the common synset \texttt{feline}, one node above where \textit{cat} belongs, thus returning $1-0.1 = 0.9$. \textit{Dog} and \textit{cat} result in $1-0.2$, as one needs to go up two nodes for both tokens to find the common synset \texttt{carnivore}. We do not go up to generic synsets, like \texttt{artifact} or \texttt{item}.

\paragraph{\textbf{Word2Vec}}
$equiv_{W2V}$ computes the cosine similarity between the vector embeddings representing the two tokens $t^i$ and $t ^j$:
\begin{align}
equiv_{W2V}(t^i, t^j) = \cos\big(emb(t^i), emb(t^j)\big),
\label{eq:w2v}
\end{align}
where $emb(t)$ is the vector representing the word embedding for the token $t$. We use the Google News word2vec models available \citep{Milokov13}\footnote{\url{https://code.google.com/archive/p/word2vec/}}. If the token is composed by more than one word (in the case of a \ac{NE} for example), their vectors are added before computing the cosine similarity. For example, \textit{car} and \textit{vehicle} obtain a cosine similarity of $0.78$, while \textit{car} and \textit{New York} result in a score of $0.07$.

\subsubsection{Full Example}

In order to illustrate the whole alignment process, consider again the seed composed of the Q/A pair \textit{Who created the telephone?/Alexander Graham Bell}, and the answer sentence \textit{Alexander Graham Bell is credited with inventing the telephone}. Matrix $M$, in Table~\ref{tab:scoresAlignmentExample}, contains:
\begin{itemize}
\item as rows, the tokens (that are not stopwords) from $Q/A$; 
\item in the columns the non-stopword tokens from $S$. 
\end{itemize}

Each cell contains the similarity score obtained by the $equiv$ functions for that pair of tokens. In the end, the preferred alignment is the one seen in the previously shown Figure~\ref{fig:rules}, chosen by the Hungarian algorithm, consisting on the cells seen in the matrix. 

% For simplicity, all \texttt{word2vec} values were ignored, as they are much lower (close to zero). (.....) A detailed analysis to the alignment method proposed can be found in Appendix~\ref{sec:alignmentExp}, where we compare its effectiveness against two state of the art aligners.

\begin{table}[t]
\centering
\caption{Scores obtained during the alignment process between the seed components: Q/A (column) and answer sentence (row). For clarity, only the best scores leading to the alignment are shown.}
\label{tab:scoresAlignmentExample}
\begin{tabular}{lcccc}
\toprule
      & Alex. ... Bell & credited & inventing & telephone \\
\midrule
created  & .. & ..    & 0.75  & ..    \\
telephone  & ..  & ..   & ..  & 1.0    \\
\midrule
Alex. ... Bell   & 1.0  & ..    & ..  & ..    \\
\bottomrule
\end{tabular}
\end{table}

Finally, before creating the pattern, the predicate-arguments of $S$ are checked for two conditions considering the alignment found. First, all tokens in $\mathcal{PA}(S)$ must belong to the alignment found, or the pattern is not created. Then, arguments can only contain tokens from either the question or the answer, but never both. These two are enforced to make sure the arguments only have information about the $Q/A$ pair and to be possible to distinguish the answer chunk from the rest of the question in the answer sentence. If the predicate-argument respects those conditions, then a pattern is created: $<Q/A, \mathcal{PA}(S), \widehat{align(Q/A, S)}>$.

Taking our running example, there are two predicate-arguments: one for \textit{credited} and another for \textit{inventing}. The first one contains the token \textit{credited} itself, which is not part of the chosen alignment, so it is discarded. For the other, all tokens in the $\mathcal{PA}$ belong to the alignment, and each argument contains tokens from either only the question or answer. Therefore, a pattern is created.

%%%%%%%%%%%%%%%%%%%%%%%%%%%%%%%%%%%%%%%%%%
%%%%%%%%%%%%%%%%%%%%%%%%%%%%%%%%%%%%%%%%%%
%%%%%%%%%%%%%%%%%%%%%%%%%%%%%%%%%%%%%%%%%%
%%%%%%%%%%%%%%%%%%%%%%%%%%%%%%%%%%%%%%%%%%

\subsection{Question Generation}
\label{sec:patternApp}

In this section we detail how \sys takes the previously learned patterns and applies them to new unseen sentences, generating new questions. In order to apply a pattern to a new sentence, the latter needs to be ``similar'' to the answer sentence originating the pattern, so that a match occurs and a question is generated. In short, a pattern contains all information necessary to transform a new sentence $S'$ into a new question $Q'$:

\begin{enumerate}
\item $\mathcal{PA}(S)$ is used to test how similar the new sentence $S'$ and $S$ are. If a match is found between $\mathcal{PA}(S)$ and a predicate-argument structure taken from $S'$, $\mathcal{PA}(S')$, then $S$ and $S'$ are considered to be similar and the pattern can be applied to $S'$;
\item If the pattern can be applied to $S'$, then $\widehat{align(Q/A, S)}$, along with $Q$, establishes how to transform $S'$ into the new $Q$-like question $Q'$.
\end{enumerate}

\subsubsection{Predicate-Argument Matching}
Let $\mathcal{P}(Q/A, S)$ = $<Q/A, \mathcal{PA}(S), \widehat{align(Q/A, S)}>$ be a pattern as previously described, and $S'$ a new unseen sentence\footnote{The same annotation process applied to $S$ in the Pattern Acquisition step is applied here to new sentences $S'$.}. \sys ``matches''  the two sentences  if there is a predicate-argument resulting from $S'$ (from now on $\mathcal{PA}(S')$) that ``matches'' $ \mathcal{PA}(S)$. Following the previous definition of predicate-arguments (predicate's root, set of arguments, and set of subtrees), let these be defined as follows:
\[\mathcal{PA}(S) =~<p_S, \{A_{S}^1, ..., A_{S}^n\}, ST(A_{S}^1) \cup ... \cup ST(A_{S}^n)>,\]
and
\[\mathcal{PA}(S') =~<p_{S'}, \{A_{S'}^1, ..., A_{S'}^m\}, ST(A_{S'}^1) \cup ... \cup ST(A_{S'}^m)>\]

The pattern $\mathcal{P}(Q/A, S)$ can only be applied to $S'$ if the following conditions are verified:
\begin{enumerate}
\item [$C_1$]: $ equiv(p_S, p_{S'}) \neq 0$; \label{enum:root}
\item [$C_2$]:  $\{A_{S}^1, ..., A_{S}^n\}$ $\subseteq$ $\{A_{S'}^1, ..., A_{S'}^m\}$;
\item [$C_3$]: $\forall st_s \in ST(A_{S}^i)~\exists st_{s'} \in ST(A_{S'}^j) : match(st_s, st_{s'}) \neq false$. \label{enum:args}
\end{enumerate}

%where $root_P$ and $root_{S'}$ are the roots of the predicate-arguments $\mathcal{PA_P}$ and $\mathcal{PA_{S'}}$, respectively, 

In other words, the predicate roots must be equivalent (Condition $C_1$), all arguments of $\mathcal{PA}(S)$ must be present in $\mathcal{PA}(S')$ (Condition $C_2$), and for each one of those arguments, the corresponding subtrees must $match$ (Condition $C_3$). If these condition are met, then $S'$ is transformed into a new question $Q'$ by replacing the tokens in $Q$ with the new tokens from $S'$, as we will see below. Function $equiv$ is the same introduced in last section for Pattern Acquisition. Function $match$ captures the equivalence between two (sub)trees, so that the system decides if a pattern should be applied or not. Two (sub)trees match if they are structurally similar and their tokens match (according to the previous $equiv$ function). We created several versions of the $match$ function, some more flexible than others. This flexibility is not only associated with the $equiv$ function, but also with the match performed over the subtrees representing parts of the sentences.

\paragraph{Strict Tree Matching}
Algorithm~\ref{alg:treeMatch} details the matching process, where $n.c$ represents the children of node $n$ (a subtree or a token). It starts by comparing the roots of the two (sub)trees (Line~\ref{l:root}). If the roots are equivalent -- using the $equiv$ function -- and the number of children is the same, the algorithm is recursively applied to their children (Line~\ref{l:forCh}). If the two trees are successfully matched recursively through their entire structure, an alignment between the two trees is returned, collected during its execution (Lines~\ref{l:align1} and~\ref{l:align2}).

\begin{algorithm}[t]
\caption{Algorithm for tree matching.}
\label{alg:treeMatch}
\begin{algorithmic}[1]
\State $match(T_1, T_2)$
\State $align \gets []$
\State $n_1 = T_1.root$
\State $n_2 = T_2.root$
\If{$equiv(n_1, n_2) \leq 0 \vee |n_1.c| \neq |n_2.c|$} \label{l:root}
	\State \Return $[]$
\Else
	\State $align \gets align(n_1, n_2)$ \label{l:align1}
\EndIf
%\If{$|n_1.c| \neq |n_2.c|$} \label{l:nCh}
%	\Return false
%\EndIf
\ForAll{$i \in n_1.c$} \label{l:forCh}
	\State $a \gets match(n_1^i, n_2^i)$
	\If{$a = []$}
		\State \Return $[]$
	\Else
		\State $align \gets a$ \label{l:align2}
	\EndIf
\EndFor
\State \Return $align$
\end{algorithmic}
\end{algorithm}

\paragraph{Subtree Matching}
This version is more relaxed than the previous one, as it accepts that arguments of $\mathcal{PA}(S')$ can have more elements in their subtrees when compared to the ones from $S$. In other words, instead of looking for a match in the subtree $ST(A_{S'}^m)$, for a given argument $A^m$, we look for a match in all subtrees of that subtree. For example, if a noun phrase is accompanied by an adjective, and the pattern only expects a noun phrase alone, \sys will be able to ignore the adjective and match the noun phrase subtree.

\paragraph{Subtree Flex Matching}
Here we are relaxing $match$ a step further. While in the above scenario we are still looking for structurally equivalent subtrees, with the \texttt{Subtree Flex Matching} we are looking to find subtrees that are just ``similar''. To do so, we use Tregrex \citep{Levy06} to create a flexible regular expression for tree nodes (for instance, \textit{N*} matches \textit{NN}). Each subtree belonging to a pattern is transformed into a template that is used in the matching process. 
For instance, in the last section we created a pattern where the predicate-argument in the pattern had its argument \texttt{A1} represented by two subtrees. Taking \texttt{NP (DT (the)) (NN (telephone))} as an example, the following expression would be generated: \texttt{/N* << ( /D* \$ /N* )}. This expression finds a subtree that starts with a noun (\texttt{N*}) that contains, as children at any level, a determinant and a noun (\texttt{D*} and \texttt{N*}, respectively). Considering again the recurring example, and a new given sentence \textit{Vasco Da Gama discovered the sea route to India}, the chunk \textit{the sea route} would be matched against \textit{the telephone}.

\paragraph{Argument Matching}
Finally, we designed a more extreme solution in which the $match$ function is true for any subtree. The idea here is that, for each argument in a pattern, the system should try to create a question by replacing the whole argument with the new one, ignoring the structure of the subtrees of either the pattern or the new sentence. 
Using \textit{Vasco da Gama} example, the whole argument \textit{the sea route to India} would match \textit{the telephone} and, thus, replace it in the original question $Q$.

In any case, if the argument being tested corresponds to the answer $A$ in the pattern, \sys checks for \acp{NE} on both subtrees being matched. If they exist, they need to be of the same type, or the generation process stops. If no \ac{NE} is found in either, then the generation process proceeds as normal. The idea is to make sure that the new question is appropriately targeting as answer the same kind of data, while not limiting \sys if the \ac{NER} fails to find a \ac{NE} (or it does not exist).

\subsubsection{Token Replacement for new Question}
Putting all together, to generate a question, if the predicate-arguments $\mathcal{PA}(S)$ and $\mathcal{PA}(S')$ match, then the new sentence $S'$ can be used to create a question, by following the alignment between $S$ and $Q/A$. The transformation of $S'$ into a new question $Q'$ (of the type $Q$) is done with the help of the alignments $align$ returned by the previous defined $match$ function. Each token aligned between $S'$ and $S$ can then be mapped to $Q$ by following the alignment between $S$ and $Q/A$ in the pattern, replacing the corresponding tokens in $Q$. In other words, each token $t^k \in S'$ that was aligned with a token $t^j \in S$ that is mapped to a token $t^i$ in the original question $Q$ will take its place in the generated question. This means, thus, that non mapped tokens in $Q$ will remain. For example, Wh-words in questions will not be mapped to tokens in $S$, being kept in the final new question and providing the same type of question.

%\[\exists \alpha(t^i, t^j) \in \widehat{align(Q, S)} \land \exists \alpha(t^j, t^k) \in align \implies replace(t^i, t^k) : t^i \in Q, t^j \in S, t^k \in S'\]

This replacement is straightforward for both \texttt{Strict Tree Matching} and \texttt{Subtree Matching}. For the other two, which are more flexible, there might not be a direct alignment between tokens in $S$ and $S'$ (the system can match longer chunks in the new sentence). Therefore, for these two approaches, for each argument matched in the predicate-argument, we replace all tokens from $Q$ which are aligned to tokens in $S$ belonging to that argument. 
%For example, \textit{the telephone} is aligned with tokens in $S$ belonging to the argument \texttt{A1}, so they will be replaced by all new tokens from $S'$ that belong to \texttt{A1}. 

Finally, we make an exception for the predicate-argument verb, which is the main verb of the question $Q$ as well. Here, we conjugate the auxiliary verb in the question in the same mode of the new matched token (from Condition 1), adjusting the main verb accordingly. This is an attempt to adjust the question formulation to the current sentence $S'$ conjugation used. For instance, if the pattern contains a question regarding a past event but the new sentence regards something yet to happen, it makes sense that the new question does not use the past sentence.

\subsubsection{Full Example}
Considering again the running example and the new given sentence \textit{Vasco Da Gama discovered the sea route to India}, because \textit{discovered} and \textit{inventing} are related ($equiv_{VB}$ function), and both have two arguments in their sentences, \texttt{A0} and \texttt{A1}, the sentences will match. Then, what remains to be tested are the subtrees that cover both arguments of each predicate. The tokens themselves will also match through semantic function $equiv$. For instance, \textit{Vasco Da Gama} and \textit{Alexander Graham Bell} are both \acp{NE} of the same type, so they are considered to match. This will result in generating different questions, depending on the tree matching strategy employed. It will not produce results using the \texttt{Strict Tree Matching}, as the subtrees representing each argument \texttt{A1} are not structurally identical, but will generate questions for the other strategies. For instance, the question \textit{Who discovered the sea route to India?/Vasco Da Gama} is generated with the \texttt{Argument Matching} strategy, by replacing the tokens in the original question with the ones from the new sentence's corresponding arguments. 

\subsection{Improving Generation with Expert's Feedback}
\label{sec:feedback}
%If \ac{QG} systems are used as an authoring tool for professors when creating content for a educational system, there is the need for the professor to manually curate those questions, not only selecting the most appropriate, but also correcting them of any mistake they might contain. This implicit feedback is never used by systems as a source of reliable data, being this way wasted. We think any correction made by a professor can play an important role in the development of a \ac{QG} system.

The main contribution of \sys relies in using the implicit feedback of the user, as we previously stated. Often, there is the need for the user to manually curate the generated questions, not only selecting the most appropriate, but also correcting them of any mistake they might contain. As a consequence of this task, every pair constituted by a sentence plus a generated question can be used by the system, after the corrections, as a new seed. This allows the system to enlarge its pool of available patterns, increasing its generation power. Nevertheless, this might also lead to a possible problem of over-generation. If a user needs to parse dozens of questions to find a good one (whatever their evaluation criteria are), then the system's usefulness might not be that interesting. Some \ac{QG} systems, such as the one described by \citet{Heilman11}, already rank the generated questions, pushing the better ones to the top, although not taking advantage of human feedback. In this section we show how we take advantage of the experts' feedback, not only to create new seeds, but also to indirectly evaluate how well the patterns are behaving. The main idea is to use the corrections made by humans as a mean to evaluate the quality of the pattern that generated the edited question. Questions needing major fixes are probably from worse patterns, while questions not requiring much editing are likely to come from well behaved patterns. Meanwhile, all generated questions with previous patterns can be used to augment the pool of available seeds. 

%This idea is closely related to the concepts of case-based reasoning introduced in the last chapter, in Section~\ref{sec:cbr}. In the last sections we described how \sys uses seeds to learn new patterns and then applies them, which corresponds to the first two steps of case-based reasoning \citep{Aamodt94}: \texttt{Retrieve} and \texttt{Reuse}. Now this section approaches the last two steps, \texttt{Revise} and \texttt{Retain}. The former is done by the user, when correcting the questions generated, while the latter is done indirectly, by scoring the patterns. 

This section is divided into two parts: the first corresponds to the validation of the generated Q/A pairs to be used by the system in a new pattern acquisition step. The second discusses the evaluation of the patterns used by the system. This part of the work borrows ideas from previous works \citep{Pantel06, Mendes13, Shima15}, discussed in the Related Work (Section~\ref{sec:evalRW}).

%\todo{Estas situações superficiais são enervantes e pouco profissionais}

\subsubsection{Learning New Seeds} 
%TheMentor \citep{Curto11}, as other systems, employed a bootstrap step to learn patterns from a set of seeds. However, past interactions are usually disregarded as a source of improvement. 

Although the task of learning new seeds can be done with no quality control at all, it might be useful to guarantee the correctness of the generated Q/A pairs (as well as the answer sentence) before adding them to the new set of seeds to be used. Having a human rating the system's output is costly, but necessary because there is no \textit{right} question to ask about a given text or sentence, but rather multiple questions can be valid. Therefore, the questions and answer sentence are presented to the user, who assesses them. Given this feedback, the system gets to know what questions were correctly generated and, thus, are a good source to be a new seed.

Each pair of new $Q/A$ and answer sentence $S$ can then be used as a new seed pair, feeding the system into a new Pattern Acquisition step. As typically the new sentences are different from the original, and the questions themselves can be edited in such a way they become different from the patterns' questions, this will lead to the creation of new patterns. These might be close to the previous patterns sometimes, but they still introduce variability to the pool of patterns, capturing this way new question formulations and enlarging the pool of available patterns.

\subsubsection{Pattern Scoring}
Given an expert in the loop we can take their feedback to score the patterns. 
%, that is, the tuple is extended to include its score $x$: $<Q, \mathcal{PA}, \widehat{align(Q, S)}, x>$. 
Let $w(\mathcal{P})$ be a function that returns the score $w_{t}$ of a pattern $\mathcal{P}$ at moment $t$. This score starts at 1.0 ($w_0 = 1$) and it is adjusted along the way, in function of the generation task, considering the previous score of the pattern, $w_{t-1}$, a new generated question $Q'$, and its edited version $Q''$ to compute a new score. This step requires two things: 
\begin{itemize}
\item a method that evaluates the difference between the generated question, $Q'$, and its edited version, $Q''$,
\item a way to incorporate that into $w^t$. 
\end{itemize}

We use lexical metrics to measure how similar both questions are. Then, to update the score, we apply the Weighted Majority Algorithm \citep{Littlestone94}, or Exponentially Weighed Average Forecast \citep{Lugosi06}. The original concept for these strategies cannot be replicated, but we adapted them to our scenario. We treat each pattern as an expert, and the generated questions as guesses from the experts. The better the guesses (that is, the more successful the generation of questions is), the better rating the expert (the pattern) will have. The successfulness of a pattern is determined by the similarity between the questions it generates and their corrected versions: 

\begin{equation}
successful(Q', Q'') = 
\begin{cases}
1 & \text{if~} sim(Q', Q'') > th\\
0 & \text{otherwise}\\
\end{cases},
\label{eq:succ}
\end{equation} 
where $th$ is a threshold and $sim$ a similarity function. For $sim$, we considered Overlap and a normalized version of \citet{Levenshtein66}. The latter gives an intuitive way to evaluate the editing effort of the human annotator in correcting a question. In specific, it gives us the edit cost considering the following operations at word level: (a) adding a new word; (b) eliminating a word; (c) transforming/replacing a word. We set the same cost for the three operations. We opted to use Levenshtein's normalized version, that takes into consideration the size of the longest question (the Levenshtein value is divided by the size of the longest question), which normalizes the scores into the range $[0$-$1]$.

The score of a pattern is updated at each step based on its previous score and the technique used, as described below.

\paragraph{Weighted Majority Algorithm} This technique penalizes a pattern if it is not successful in the generated questions and rewards it otherwise:

\begin{equation}
w(\mathcal{P})_{(Q', Q'')} =
\begin{cases}
1 & \text{if  t = 0}\\
w_{t-1}(1-b)^{-1} & \text{if \textit{successful}(Q', Q'')}\\
w_{t-1}(1-l) & \text{otherwise}\\
\end{cases}
\label{eq:pesos}
\end{equation}
where $l$ is the loss penalty for a non successful generation and $b$ the bonus reward for a successful generation. If $b$ is set to 0, then no bonus is awarded, becoming the score a decaying factor only.

\paragraph{Exponentially Weighed Average Forecast} Here the score of a pattern is updated by a decaying factor relative to its successfulness, i.e., the better it performs, the less its score is penalized:

\begin{equation}
w(\mathcal{P})_{(Q', Q'')} =
\begin{cases}
1 & \text{if  t = 0}\\
w_{t-1}*e^{-l *\frac{1}{sim(Q', Q'')}} & \text{otherwise}\\
\end{cases}
\label{eq:pesos1}
\end{equation}
where $l$ is the loss penalty and $sim$ is the similarity function presented before. 

This process is applied to each question generated by each pattern at each time.

%%%%%%%%%%%%%%%%%%%%%%%%%%%%%%
%%%%%%%%%%%%%%%%%%%%%%%%%%%%%%
%%%%%%%%%%%%%%%%%%%%%%%%%%%%%%
%%%%%%%%%%%%%%%%%%%%%%%%%%%%%%
%%%%%%%%%%%%%%%%%%%%%%%%%%%%%%

\section{Experimental Setup}\label{sec:es}

In this section we describe our experimental setup, as well as the metrics used in the evaluation process. 

\subsection{Evaluation}
To compare \sys with state of the art systems, before incorporating the implicit feedback and study its impact, we used two \ac{QG} systems covering two different approaches to the problem. The first one is the work of \citet{Heilman09}, from now on H\&S, and employs a rule based approach. The second one is the work of \citet{Du17}, from now on D\&A, and is a neural network system.  We set up H\&S to prefer Wh-questions and limit questions to 30 tokens (\textsl{--prefer-wh --max-length 30}), and retrieved the top questions with a score above 2.5. Considering \du, we trained the network with the same configuration as the authors. The model generates a question for each input sentence, in both the sentence- and paragraph-level models. We chose the sentence-level model to keep all systems equal, and, also, to mimic the annotators who only had access to the individual sentences.  To bootstrap \sys, we used 8 seeds (Table \ref{tab:seeds}), as suggested by~\citet{Rodrigues18}. We also use all tree matching strategies together, reporting overall results only.

\begin{table*}[t]
\center
\caption{Seeds used in the Pattern Acquisition phase of \sys.}
\label{tab:seeds}
\scriptsize
\begin{tabular}{p{7cm} p{5cm}}
\toprule
\multicolumn{1}{c}{\textbf{Support Sentence}} & \multicolumn{1}{c}{\textbf{Question}}\\
\midrule
Leonardo da Vinci was born on April 15, 1452. & When was Leonardo da Vinci born?\\
Lee Harvey Oswald was assassinated by Jack Ruby. & Who killed Lee Harvey Oswald? \\ 
 Paris is located in France. & Where is Paris located?\\
Porto is located 313 km from Lisbon. & How far is Lisbon from Porto?\\
Yesterday, Bob took butter from the fridge. & Where did Bob take butter from?\\
John baked cookies in the oven. & What did John bake in the oven?\\
Cooking is the art, technology, science and craft of preparing food for consumption. & What is cooking?\\
Science is a systematic enterprise that builds and organizes knowledge in the form of testable explanations and predictions about the universe. & What is a systematic enterprise that builds and organizes knowledge in the form of testable explanations and predictions about the universe?\\
\bottomrule
\end{tabular}
\end{table*}

As for the evaluation process, we used a publicly available library containing different lexical and semantic metrics. The Maluuba project \citep{Sharma17}\footnote{\url{https://github.com/Maluuba/nlg-eval}} contains lexical metrics typically used, like BLEU, METEOR, and ROUGE, and other metrics based on word embeddings: \ac{EACS}, \ac{STCS}, \ac{VECS}, and \ac{GMS}:
\begin{itemize}
\item BLEU \citep{Papineni02} is typically used to evaluate machine translation systems and has also been used for \ac{QG}, as it compares a candidate sentence with a reference of acceptable hypothesis. The BLEU score is computed by calculating a modified precision on the shared $n$-grams between the candidate and the reference. The values employed for $n$ are typically between 1 and 4. 
\item METEOR \citep{Banerjee05} is a metric that is supposed to correlate better with human evaluations. It aligns the candidate sentence with a reference sentence by performing token alignment. The precision and recall of those alignments are used to compute a F-score that will lead to the final score. 
\item ROUGE \citep{Lin04} is another lexical metric that also computes a F-score between the candidate sentence and the reference hypothesis. In the Maluuba library ROUGE\_L is used, which is based on the Longest Common Subsequence between the two sentences. 
\item \ac{EACS} computes the cosine similarity of two sentence embeddings. The sentence embedding is formed by averaging the word embeddings of each of the sentences' tokens. 
\item \ac{STCS} also computes the cosine similarity between two sentence embeddings. However, it is based on the Skip-Thought model \citep{Kiros15}, a recurrent network trained to predict the next and previous sentence of the input sentence, which is encoded into a sentence embedding. These embeddings showed to have good performance in semantic relatedness, and are used here as an alternative to averaging the sentence's token embeddings.
\item \ac{VECS} \citep{Forgues14} also computes the cosine similarity between two sentence embeddings, but in this case each embedding is created by taking the most extreme (maximum) value among all token embeddings, for each dimension. 
\item \ac{GMS} is the only embedding-based score that does not use sentence embeddings. Instead, it takes each token embedding in the candidate and maximizes its similarity with a token on the reference, summing all those scores for all tokens. Then it performs the same task inverting the candidate and reference hypothesis roles, and averages both scores \citep{Rus12b}.
\end{itemize}

\subsection{Obtaining Feedback}

In order to capture the effects of the implicit feedback over time, we simulate the iteration process of system-user interaction by batching the input source. Therefore, instead of using a single target corpus as a one-time input, we batch it in multiple smaller inputs, allowing the system to evolve over time by using feedback obtained in previous batches. For this purpose, one of the authors, playing the teacher’s role, corrected the generated questions, or discarded them if no reasonable fix could be found, for each batch. Questions corrected to a different type would still be used as new seeds, but were considered as discarded for weighing purposes.

Algorithm~\ref{alg:iterative} shows the pseudo-code for the iterative process. Being given an original set of seeds, patterns are created as previously described in Section~\ref{sec:patternAcq} (line 5). The learning materials are divided in batches of sentences to which patterns are applied, resulting in new questions (line 8). Every generated question is presented to the user to be corrected or discarded (line 10). After being corrected by the user, each question is associated with the sentence that originated it and added to the set of seeds, allowing the creation of new patterns (line 12), and also used to score the pattern that generated it (line 14). Finally, the patterns that generated only discarded questions are removed from the pool of patterns, reducing the number of ill questions to be generated in future batches (line 16). The process repeats for all batches (lines 4-18).

\begin{algorithm}[t]
\begin{algorithmic}[1]
\caption{Iterative loop for online learning with \sys.}\label{increasing patterns}
\label{alg:iterative}
\State $seeds \gets \{(s_1, q_1), ..., (s_n, q_n)\}$
\State $Batches \gets \{Batch_1, ..., Batch_m\}$
\State $P \gets \{\}$
\ForAll {$b \in Batches$} 
  \State $P \gets P \cup generatePatterns(seeds)$
  \State $seeds \gets \{\}$
  \ForAll {$s \in B$} // $s$ is a single sentence in the batch $b$
    \State $Q \gets generateQuestions(s, P)$
    \ForAll {$q \in Q$}
       \State $q' \gets correctQuestion(q)$
       \If{$q' \neq null$}
         \State $seeds \gets seeds \cup \{(s, q')\}$
       \EndIf 
       \State $P \gets weighPattern(q, q', P)$
     \EndFor
     \State $P \gets discardPatterns(P)$
  \EndFor
\EndFor
\end{algorithmic}
\end{algorithm}

As each pattern leads to many questions, that might be discarded or not, it is not trivial to illustrate the end-to-end process on how new patterns are generated and the new question's scores are calculated. However, in Table~\ref{tab:iterationExample}, we demonstrate a case from the first to the second iteration of one of the experimental runs. In the table we can see on the last column what happened to the generated question and how that influences the pattern's score and, thus, the score of the questions generated in the next iteration, pushing a good question to the top.

\begin{table}[t!]
\centering
\caption{Example of questions' contributions (among many others) to the evolution of pattern scores, from the first to the second iteration of one of the experimental runs.}
\label{tab:iterationExample}
\scriptsize
\begin{tabular}{rp{2cm}p{3cm}cp{2cm}p{2cm}}
\toprule
\textbf{Score} & \textbf{Question} & \textbf{Sentence} & \textbf{Matching} & \textbf{Original Pattern Q}  & \textbf{Result}\\
\midrule
0.0	& What did de Visme build? & The effective acquisition of the property took place in 1863, with the architect James Knowles beginning the work of transforming what remained of the house built by de Visme.  & FLEX\_SUBTREE & What did Kafka write? & \paragraph{} Question kept \paragraph{} New pattern \paragraph{} Leading to better score of original pattern \\
0.0	& When was the surrounding gardens had? & Over the years, the surrounding gardens have welcomed plant species from all over the world. & FLEX\_SUBTREE  & When was Leonardo da Vinci born? & \paragraph{} Question discarded \paragraph{} No new pattern \paragraph{} Leading to lower score of original pattern\\
\midrule
2.75 &	What did the Portuguese State buy? & The estate and the Palace were bought by the Portuguese State in 1949. & FLEX\_SUBTREE & What did Kafka write? & -\\
0.67 &	When was The Park Palace Monserrate been? & The Park and Palace of Monserrate were classified as a Property of Public Interest in 1993, and were included in the Cultural Landscape of Sintra, which has been classified by UNESCO as World Heritage since 1995. & FLEX\_SUBTREE  & When was Leonardo da Vinci born? & -\\
\bottomrule
\end{tabular}
\end{table}

\subsection{Corpora}
We used \McQGE in these experiments \citep{Rodrigues21}, as it is an extensive reference that allows an easier automatic validation of the results. It contains 73 source sentences to generate questions from, and over 2k questions as reference. The batches for Algorithm~\ref{alg:iterative} were created by splitting the corpus in equal parts. We tried three different sizes: 7 (leading to roughly 10 batches), 10 (7 batches), and 12 (6 batches). Note that batches should neither be too large (user's experience), nor too small (many batches lead to smaller learning increments).

To evaluate the generated questions against the reference we use the metrics presented before, but at top $N$, i.e., $N$ is a cut to the top questions in the list of all questions. In these experiments we use $N=5$, $N=10$ and $N=20$, as if the user would be only presented those top $N$ questions.

\subsection{Pattern Scoring and Baselines}
From Section~\ref{sec:feedback}, we mentioned two techniques to update the score of a given pattern: \ac{WMA} and \ac{EWAF}. Each has specific values that need to be set, like the loss penalty. In either case, they require $successful$ function to dictate if a pattern was successful, which relies in a similarity function $sim$. We run different parameterizations for the weighing techniques described. Besides the different $sim$ functions described before (Section~\ref{sec:feedback}), we also set the $penalties$ and $bonus$ for both. As both strategies were adapted to our problem, we empirically chose these values. For \ac{WMA} we set the following weights for $penalty$ and $bonus$ parameters, respectively, to: $0.1$, $0.2$, and $0.1$, $0.3$, $0.5$. The threshold $th$ (Equation~\ref{eq:succ}) for $sim$ function was set to $0.9$ and $0.8$. For \ac{EWAF} we set the $penalty$ to values of $0.1$ and $0.2$ -- the threshold $th$ and $bonus$ parameters are not applicable. Table~\ref{tab:params} summarizes this information. 

\begin{table}[t!]
\centering
\caption{Parametrization of the different variables in the weighing strategies, \ac{WMA} and \ac{EWAF}: function $sim$, its threshold $th$, penalties and bonus values.}
\label{tab:params}
\begin{tabular}{lcccc}
\toprule
& $\bm{sim}$ & $\bm{th}$ & $\bm{penalty}$ & $\bm{bonus}$\\
\midrule
\ac{WMA} & Overlap, Lev & 0.9, 0.8 & 0.1, 0.2 & 0.1, 0.3, 0.5\\
\ac{EWAF} & Overlap, Lev & - & 0.1, 0.2 & -\\
\bottomrule
\end{tabular}
\end{table}

We also set two baselines. The first (\texttt{original patterns}) corresponds to the original patterns applied to all batches (i.e., there is no learning of patterns with new batches). Because there is no ranking of the generated questions, and to get a more accurate baseline, we average three different random orderings for this baseline. The second \texttt{baseline} corresponds to the algorithm of learning patterns from new seeds, but with no weighing strategy in place. Again, the generated questions in each batch are not ordered, so all the reported results correspond to the average of three different random orderings as well.

%\subsection{Iteration Example}

%%%%%%%%%%%%%%%%%%%%%%%%%
%%%%%%%%%%%%%%%%%%%%%%%%%
%%%%%%%%%%%%%%%%%%%%%%%%%
%%%%%%%%%%%%%%%%%%%%%%%%%

\section{Results}\label{sec:results}
In this section we report the results obtained by \sys, comparing first its non-learning version with two state of the art systems, and then how it improves using the strategies and setup discussed in last section.

\subsection{Without Using Implicit Feedback}

Being given the 73 sentences from the \McQGE corpus, H\&S system generated 108 questions (after discarding more than 200 questions below the 2.5 score threshold), and \sys 209 questions. As \du is limited to one question per sentence, 73 questions were generated. Table~\ref{tab:resultsMcQGE} shows the obtained results. 

\begin{table}[t!]
\centering
\caption{Overall scores obtained with automatic metrics on \McQGE, for H\&S, D\&A
and \sys.}
\label{tab:resultsMcQGE}
\small
\begin{tabular}{lrrrrrrrr}
\toprule
& ROUGE & METEOR & BLEU1 & BLEU4 & EACS & GMS & STCS & VECS\\
\midrule
H\&S & \textbf{69.00}  & 46.38  & 83.71 & \textbf{45.56}  & 92.51 & \textbf{86.11} & 73.26 & 77.92 \\
%\midrule
\du & 63.71  & 37.58  & 77.40 & 26.63 &  \textbf{92.52} & 85.51 &  \textbf{74.47} & 77.54\\
\midrule
%\sys Strict & 50.00 & 23.02 &  75.00 & 1.8e-8 & 83.39 & 70.11 & 51.87 & 67.58\\
\sys SubT & 60.60 & 30.67 & \textbf{95.93} & 28.45 & 86.40 & 80.38 & 60.28 & 74.07\\
\sys  FlexSubT & 64.66 & 40.63 & 86.91 & 35.25 & 90.10 & 84.14 & 67.09 & 77.96\\
\sys Args & 65.81 & \textbf{46.44} & 81.80 & 40.61 & 92.25 & 85.86 & 71.17 & \textbf{80.89}\\
\sys All & 63.13 & 41.09 & 77.90 & 34.33 & 90.52 & 83.97  & 67.60& {77.90}\\
\bottomrule
\end{tabular}
\end{table}

Results show \sys performs overall better than D\&A, but without surpassing H\&S, which has the best overall results. Some configurations of \sys surpass H\&S in some metrics (highlighted), but all together \sys does not surpass H\&S. We can also observe that semantic metrics are less prone to distinguish the systems as lexical ones. 

\subsection{Using Implicit Feedback}

In this section we study how \sys benefits from incorporating implicit feedback, and use all of \sys's configurations together. The combination of all different parameters from Table~\ref{tab:params} leads to a large number of possible combinations. After inspection of the results obtained, we concluded that some configurations tend to perform similarly, if not equally to others. Therefore, we omitted most configurations from the tables and discussion, for sake of readability, presenting the overall best results, analyzed at large, i.e., the configurations that consistently showed the best results across all runs: \texttt{EWAF-Overlap-01} and \texttt{WMA-Lev-08-02}. The names indicate the configuration following the same order of Table~\ref{tab:params}. For instance, \texttt{WMA-Lev-08-02} corresponds to using \ac{WMA} as weighing strategy, with $sim$ function being Levenshtein, the threshold $th$ $0.8$, and the $penalty$ $0.2$. 

%Complete data can be found in Appendix~\ref{sec:appExpert}, where all results are listed.

\begin{table}[t!]
\centering
\caption{Comparison of the weighing strategies against the baselines, measured by automatic metrics on \McQGE, at top 5, 10, and 20. Scores normalized by the best score obtained in each metric. Overall results for batches of size 10 (7 batches) -- averaged on all but first batch.}
\label{tab:resultsIter10-5}
\small
\begin{tabular}{lrrrrrrrr}
\toprule
@5 & ROUGE & METEOR & BLEU1 & BLEU4 & EACS & GMS & STCS & VECS\\
\midrule
\midrule
Original Patterns  & 0.74 & 0.62 & 0.83 & 0.48 & 0.94 & 0.84 & 0.76 & 0.82\\
\midrule  
%\midrule
Baseline  & 0.92 & 0.81 & \textbf{1.00} & 0.67 & 0.99 & 0.91 & 0.86 & \textbf{1.00} \\
\midrule 
\midrule  
%EWAF-Lev-01  & 0.97 & \textit{0.78} & \textit{0.99} & 0.80 & 0.99 & 0.91 & 0.89 & \textit{0.99} \\
%EWAF-Lev-02  & 0.99 & 0.93 & \textit{0.96} & 0.97 & \textbf{1.00} & \textbf{1.00} & 0.95 & \textit{0.99} \\
%\midrule  
EWAF-Overlap-01  & \textbf{1.00} & \textbf{1.00} & \textit{0.96} & \textbf{1.00} & \textbf{1.00} & \textbf{1.00} & \textbf{0.96} & \textit{0.99} \\
%EWAF-Overlap-02  & 0.97 & 0.82 & \textit{0.99} & 0.87 & 0.99 & 0.91 & 0.88 & \textit{0.99} \\
%\midrule  
\midrule 
%WMA-Lev-08-01  & 0.97 & \textit{0.79} & \textit{0.98} & 0.85 & {0.99} & 0.91 & \textbf{0.93} & \textit{0.98} \\
WMA-Lev-08-02  & 0.97 & \textbf{0.82} & \textit{0.99} & 0.88 & {0.99} & 0.91 & \textbf{0.93} & \textit{0.99} \\
%WMA-Lev-09-01  & \textbf{0.98} & 0.81 & \textit{0.99} & \textbf{0.89} & {0.99} & \textbf{0.92} & \textbf{0.93} & \textbf{1.00} \\
%WMA-Lev-09-02  & 0.97 & \textit{0.79} & \textit{0.98} & 0.85 & {0.99} & 0.91 & \textbf{0.93} & \textit{0.99} \\
%\midrule 
%\midrule 
%WMA-Overlap-08-01-B03  & \textit{0.90} & \textbf{0.85} & \textit{0.92} & \textit{0.45} & {0.99} & \textbf{0.92} & \textbf{1.00} & \textit{0.97} \\
\midrule
\midrule
@10 & ROUGE & METEOR & BLEU1 & BLEU4 & EACS & GMS & STCS & VECS\\
\midrule
\midrule
Original Patterns  & 0.74 & 0.68 & 0.82 & 0.53 & 0.95 & 0.84 & 0.78 & 0.83 \\
\midrule  
%\midrule  
Baseline  & 0.94 & 0.93 & \textbf{1.00} & 0.83 & 0.99 & 0.92 & 0.90 & \textbf{1.00} \\
\midrule  
\midrule  
%EWAF-Lev-01  & 0.97 & 0.94 & \textit{0.98} & 0.85 & 0.99 & 0.92 & 0.92 & \textit{0.98} \\
%EWAF-Lev-02  & 0.98 & 0.94 & \textit{0.99} & 0.96 & \textbf{1.00} & \textbf{1.00} & 0.95 & \textit{0.98} \\
%\midrule  
EWAF-Overlap-01  & \textbf{1.00} & \textbf{0.99} & \textit{0.99} & \textbf{1.00} & \textbf{1.00} & \textbf{1.00} & \textbf{0.96} & \textit{0.99} \\
%EWAF-Overlap-02  & 0.97 & 0.96 & \textit{0.98} & 0.88 & \textbf{1.00} & 0.92 & 0.92 & \textit{0.98} \\
%\midrule  
\midrule  
%WMA-Lev-08-01  & 0.97 & 0.98 & \textit{0.98} & 0.90 & {0.99} & {0.92} & 0.96 & \textit{0.98} \\
WMA-Lev-08-02  & \textbf{0.98} & \textbf{1.00} & \textit{0.98} & 0.92 & {0.99} & {0.92} & \textbf{0.97} & \textbf{1.00} \\
\midrule
\midrule
@20 & ROUGE & METEOR & BLEU1 & BLEU4 & EACS & GMS & STCS & VECS\\
\midrule
\midrule
Original Patterns & 0.76 & 0.75 & 0.81 & 0.65 & 0.96 & 0.85 & 0.85 & 0.85 \\
\midrule 
%\midrule 
Baseline & 0.89 & 0.93 & 0.96 & 0.87 & 0.98 & 0.91 & 0.97 & 0.95 \\
\midrule 
\midrule 
%EWAF-Lev-01 & 0.92 & 0.97 & \textit{0.95} & 0.90 & 0.98 & 0.92 & 0.99 & 0.98 \\
%EWAF-Lev-02 & 0.99 & 0.97 & \textbf{1.00} & 0.96 & \textbf{1.00} & \textbf{1.00} & 0.99 & \textbf{1.00} \\
%\midrule 
EWAF-Overlap-01 & \textbf{1.00} & \textbf{1.00} & \textbf{1.00} & \textbf{1.00} & \textbf{1.00} & \textbf{1.00} & \textbf{1.00} & \textbf{1.00} \\
%EWAF-Overlap-02 & 0.92 & 0.97 & \textit{0.95} & 0.90 & 0.98 & 0.92 & 0.99 & 0.98 \\
%\midrule 
\midrule 
%WMA-Lev-08-01 & 0.90 & 0.94 & \textit{0.94} & 0.83 & 0.98 & 0.91 & \textbf{1.00} & 0.97 \\
WMA-Lev-08-02 & 0.90 & 0.94 & \textit{0.94} & 0.83 & 0.98 & 0.91 & \textbf{1.00} & 0.97 \\
\bottomrule
\end{tabular}
\end{table}

The results are normalized along each column by the best score obtained. We opted to present results this way so it is easier to understand to what degree strategies can improve over others. A score of $1.00$ thus represents the best obtained score for that metric among all strategies, including both baselines. Because we omitted many configurations, it is possible that a $1.00$ score is not shown in the tables for some metrics, but note that this is an exception, not the norm. 

In the following tables, highlighted results correspond to improvements against the baseline for that strategy, but not necessarily the best result attained overall (which will always be $1.00$). Italicized values correspond to results worse than the baseline. Finally, non-highlighted results are better than the baseline, but not the best results for that strategy. For instance, in Table~\ref{tab:resultsIter10-5}, the fourth row corresponds to the results obtained using configuration \texttt{WMA-Lev-08-02} for top 5 questions, and it surpasses the baseline for all metrics except BLEU1 and \ac{VECS} (in italic), although it does not surpass \texttt{EWAF-Overlap-01} in any metric, being the best of all \ac{WMA} configurations (not shown) on METEOR and \ac{STCS}, reason why those results are highlighted in that row.

The first iteration was run for batches of size 10 (7 batches). Table~\ref{tab:resultsIter10-5} shows the results for the top $N$ questions ranked, with $N$ equal to 5, 10, and 20, and for the two configurations selected, averaging the results obtained in all but the first batch, as the learning phase only starts after acquiring data from the first batch.

Analyzing the results per strategy for all cuts of $N$, the first thing to note is that the \texttt{baseline} improves over the \texttt{original patterns} (from 17\% up for lexical measures), which means that learning new seeds alone already improves the original \sys. When considering the weighing strategies, results were improved even more, with \texttt{EWAF-Overlap-01} being the more consistent strategy, improving in different proportions depending on the metric, $N$, and baselines considered.

\begin{table}[t!]
\centering
\caption{Comparison of the weighing strategies against the baselines, measured by automatic metrics on \McQGE, at top 5, 10, and 20. Scores normalized by the best score obtained in each metric. Overall results for batches of size 7 (10 batches) -- averaged on all but first batch.}
\label{tab:resultsIter7-5}
\small
\begin{tabular}{lrrrrrrrr}
\toprule
@5 & ROUGE & METEOR & BLEU1 & BLEU4 & EACS & GMS & STCS & VECS\\
\midrule
\midrule
Original Patterns  & 0.85 & 0.73 & 0.85 & 0.52 & 0.95 & 0.93 & 0.87 & 0.85 \\
\midrule
%\midrule
Baseline  & 0.95 & 0.91 & \textbf{1.00} & 0.68 & \textbf{1.00} & \textbf{1.00} & 0.97 & 0.99 \\
\midrule
\midrule
%EWAF-Lev-01  & \textbf{1.00} & \textbf{1.00} & \textbf{1.00} & \textbf{1.00} & \textbf{1.00} & \textbf{1.00} & \textbf{0.98} & \textbf{1.00} \\
%EWAF-Lev-02  & \textbf{1.00} & \textbf{1.00} & \textbf{1.00} & \textbf{1.00} & \textbf{1.00} & \textbf{1.00} & \textbf{0.98} & \textbf{1.00} \\
%\midrule
EWAF-Overlap-01  & \textbf{1.00} & \textbf{1.00} & \textbf{1.00} & \textbf{1.00} & \textbf{1.00} & \textbf{1.00} & \textbf{0.98} & \textbf{1.00} \\
%EWAF-Overlap-02  & 0.99 & \textbf{1.00} & \textit{0.93} & 0.99 & \textbf{1.00} & \textbf{1.00} & \textbf{0.98} & \textbf{1.00} \\
%\midrule
\midrule
%WMA-Lev-08-01  & 0.95 & 0.94 & \textit{0.97} & 0.91 & \textit{0.99} & \textit{0.98} & \textbf{1.00} & 0.99 \\
WMA-Lev-08-02  & 0.95 & \textbf{0.97} & \textit{0.97} & \textbf{0.92} & \textit{0.99} & \textit{0.98} & \textbf{1.00} & \textit{0.96} \\
%WMA-Lev-09-01  & \textbf{0.96} & 0.94 & \textit{0.98} & 0.90 & \textit{0.99} & \textit{0.98} & 0.99 & \textit{0.96} \\
%WMA-Lev-09-02  & 0.95 & 0.94 & \textit{0.97} & 0.91 & \textit{0.99} & \textit{0.98} & \textbf{1.00} & \textit{0.96} \\
%\midrule
%\midrule
%WMA-Overlap-08-01-B03  & \textit{0.90} & \textit{0.82} & \textit{0.99} & \textit{0.59} & \textbf{1.00} & \textit{0.99} & \textit{0.84} & \textit{0.96} \\
\midrule
\midrule
@10 & ROUGE & METEOR & BLEU1 & BLEU4 & EACS & GMS & STCS & VECS\\
\midrule
\midrule
Original Patterns  & 0.99 & 0.87 & 0.89 & \textbf{1.00} & 0.97 & 0.97 & 0.89 & 0.91 \\
\midrule 
%\midrule 
Baseline  & \textit{0.91} & 0.88 & 0.95 & \textit{0.71} & 0.98 & 0.98 & 0.95 & 0.97 \\
\midrule 
\midrule 
%EWAF-Lev-01  & \textbf{1.00} & \textbf{1.00} & \textbf{1.00} & \textit{0.92} & \textbf{1.00} & \textbf{1.00} & 0.98 & \textbf{1.00} \\
%EWAF-Lev-02  & \textbf{1.00} & \textbf{1.00} & \textbf{1.00} & \textit{0.94} & \textbf{1.00} & \textbf{1.00} & 0.98 & \textbf{1.00} \\
%\midrule 
EWAF-Overlap-01  & \textbf{1.00} & \textbf{1.00} & \textbf{1.00} & \textit{0.94} & \textbf{1.00} & \textbf{1.00} & 0.98 & \textbf{1.00} \\
%EWAF-Overlap-02  & \textbf{1.00} & \textbf{1.00} & 0.98 & \textit{0.94} & \textbf{1.00} & \textbf{1.00} & 0.98 & \textbf{1.00} \\
%\midrule 
\midrule 
%WMA-Lev-08-01  & 0.98 & \textbf{0.96} & 0.99 & \textit{0.88} & \textbf{1.00} & \textbf{1.00} & 0.98 & \textbf{1.00} \\
WMA-Lev-08-02  & 0.98 & \textbf{0.96} & 0.99 & \textit{0.88} & \textbf{1.00} & \textbf{1.00} & 0.98 & 0.97 \\
\midrule
\midrule
@20 & ROUGE & METEOR & BLEU1 & BLEU4 & EACS & GMS & STCS & VECS\\
\midrule
\midrule
Original Patterns  & 0.91 & 0.85 & 0.86 & 0.88 & 0.94 & 0.91 & 0.85 & 0.87 \\
\midrule 
%\midrule 
Baseline  & 0.93 & 0.92 & 0.95 & \textit{0.82} & 0.99 & 0.98 & 0.97 & 0.97 \\
\midrule 
\midrule 
%EWAF-Lev-01  & 0.97 & 0.98 & 0.97 & 0.88 & \textbf{1.00} & 0.99 & 0.99 & 0.99 \\
%EWAF-Lev-02  & \textbf{1.00} & \textbf{1.00} & 0.99 & 0.99 & \textbf{1.00} & \textbf{1.00} & \textbf{1.00} & \textbf{1.00} \\
%\midrule 
EWAF-Overlap-01  & \textbf{1.00} & \textbf{1.00} & \textbf{1.00} & \textbf{1.00} & \textbf{1.00} & \textbf{1.00} & 0.99 & \textbf{1.00} \\
%EWAF-Overlap-02  & 0.97 & 0.97 & 0.98 & 0.89 & \textbf{1.00} & 0.99 & 0.99 & 0.99 \\
%\midrule 
\midrule 
%WMA-Lev-08-01  & \textbf{0.99} & \textbf{0.98} & \textbf{0.98} & \textbf{0.91} & \textbf{1.00} & \textbf{1.00} & \textbf{1.00} & \textbf{1.00} \\
WMA-Lev-08-02  & \textbf{0.99} & \textbf{0.98} & \textbf{0.98} & \textbf{0.91} & \textbf{1.00} & \textbf{1.00} & \textbf{1.00} & \textbf{1.00} \\
\bottomrule
\end{tabular}
\end{table}

Table~\ref{tab:resultsIter7-5} shows the same evaluation procedure for the top 5, 10 and top 20 ranked questions for the same experiment with batches of size 7 (10 batches). For $N = 5$ results for \texttt{WMA} show few gains compared to the baselines, while the \texttt{EWAF} configuration still surpasses them. For greater values of $N$, the pattern witnessed in the previous experiment (7 batches) occurs as well: \texttt{EWAF} strategy shows the best improvements, but \texttt{WMA} also shows improvements across all metrics.

Finally, we did the same experiment for larger batches of 12 sentences each, leading to 6 batches. Table~\ref{tab:resultsIter12-5} shows the results obtained with the two configurations. The trend previously seen also applies here, with the most improvements being witnessed at top 5 and top 10 across all metrics, specially for \texttt{EWAF} configurations. For top 20, improvements are less pronounced, while the \texttt{baseline} still improves over the \texttt{original patterns}.

\begin{table}[t!]
\centering
\caption{Comparison of the weighing strategies against the baselines, measured by automatic metrics on \McQGE, at top 5, 10, and 20. Scores normalized by the best score obtained in each metric. Overall results for batches of size 12 (6 batches) -- averaged on all but first batch.}
\label{tab:resultsIter12-5}
\small
\begin{tabular}{lrrrrrrrr}
\toprule
@5 & ROUGE & METEOR & BLEU1 & BLEU4 & EACS & GMS & STCS & VECS\\
\midrule
\midrule
Original Patterns  & 0.84 & 0.79 & 0.90 & 0.66 & 0.94 & 0.91 & 0.80 & 0.90 \\
\midrule 
Baseline  & 0.93 & 0.89 & 0.93 & 0.88 & 0.97 & 0.95 & 0.90 & 0.94 \\
\midrule 
\midrule 
%EWAF-Lev-01  & \textbf{1.00} & \textbf{1.00} & \textbf{0.94} & \textbf{1.00} & 0.97 & 0.95 & \textbf{1.00} & 0.95 \\
%EWAF-Lev-02  & \textbf{1.00} & \textbf{1.00} & \textbf{0.94} & \textbf{1.00} & 0.97 & 0.95 & \textbf{1.00} & \textbf{0.99} \\
%\midrule 
EWAF-Overlap-01  & \textbf{1.00} & \textbf{1.00} & \textbf{0.94} & \textbf{1.00} & 0.97 & 0.95 & \textbf{1.00} & \textbf{0.99} \\
%EWAF-Overlap-02  & \textbf{1.00} & \textbf{1.00} & \textbf{0.94} & \textbf{1.00} & 0.97 & 0.95 & \textbf{1.00} & 0.95 \\
%\midrule 
\midrule 
%WMA-Lev-08-01  & \textbf{0.96} & \textbf{0.90} & \textit{0.91} & \textit{0.85} & \textit{0.96} & \textit{0.94} & \textbf{0.98} & \textit{0.91} \\
WMA-Lev-08-02  & \textbf{0.96} & \textbf{0.90} & \textit{0.91} & \textit{0.85} & \textit{0.96} & \textit{0.94} & \textbf{0.98} & \textbf{0.96} \\
%WMA-Lev-09-01  & 0.95 & 0.89 & \textit{0.89} & \textit{0.78} & \textit{0.96} & \textit{0.94} & \textbf{0.98} & \textbf{0.96} \\
%WMA-Lev-09-02  & 0.95 & 0.89 & \textit{0.89} & \textit{0.78} & \textit{0.96} & \textit{0.94} & \textbf{0.98} & \textbf{0.96} \\
%\midrule 
%\midrule 
%WMA-Overlap-08-01-B03  & \textbf{0.99} & \textbf{0.98} & \textbf{1.00} & \textit{0.77} & \textbf{1.00} & \textbf{1.00} & \textit{0.86} & \textbf{1.00} \\
\midrule
\midrule
@10 & ROUGE & METEOR & BLEU1 & BLEU4 & EACS & GMS & STCS & VECS\\
\midrule
\midrule
Original Patterns  & 0.90 & 0.86 & 0.94 & 0.75 & 0.96 & 0.94 & 0.87 & 0.92 \\
\midrule 
%\midrule 
Baseline  & 0.94 & 0.92 & 0.97 & 0.81 & 0.98 & 0.97 & 0.95 & 0.98 \\
\midrule 
\midrule 
%EWAF-Lev-01  & \textbf{0.99} & \textbf{1.00} & \textit{0.95} & \textbf{1.00} & 0.98 & 0.97 & \textbf{1.00} & 0.96 \\
%EWAF-Lev-02  & \textbf{0.99} & \textbf{1.00} & \textit{0.95} & \textbf{1.00} & 0.98 & 0.97 & \textbf{1.00} & \textbf{1.00} \\
%\midrule 
EWAF-Overlap-01  & \textbf{0.99} & \textbf{1.00} & \textit{0.95} & \textbf{1.00} & 0.98 & 0.97 & \textbf{1.00} & \textbf{1.00} \\
%EWAF-Overlap-02  & \textbf{0.99} & \textbf{1.00} & \textit{0.95} & \textbf{1.00} & 0.98 & 0.97 & \textbf{1.00} & 0.96 \\
%\midrule 
\midrule 
%WMA-Lev-08-01  & \textbf{1.00} & \textbf{0.99} & \textit{0.95} & \textbf{0.95} & 0.98 & \textbf{0.98} & \textbf{1.00} & 0.96 \\
WMA-Lev-08-02  & \textbf{1.00} & \textbf{0.99} & \textit{0.95} & \textbf{0.95} & 0.98 & \textbf{0.98} & \textbf{1.00} & \textbf{1.00} \\
\midrule
\midrule
@20 & ROUGE & METEOR & BLEU1 & BLEU4 & EACS & GMS & STCS & VECS\\
\midrule
\midrule
Original Patterns  & 0.88 & 0.81 & 0.90 & 0.74 & 0.98 & 0.94 & 0.88 & 0.89 \\
\midrule 
%\midrule 
Baseline  & 0.97 & 0.93 & \textbf{1.00} & 0.87 & 0.99 & 0.99 & \textbf{1.00} & 0.98 \\
\midrule 
\midrule 
%EWAF-Lev-01  & 0.99 & \textbf{0.99} & \textit{0.97} & \textbf{0.97} & \textbf{1.00} & \textbf{1.00} & \textit{0.97} & \textbf{1.00} \\
%EWAF-Lev-02  & 0.99 & \textbf{0.99} & \textit{0.98} & \textbf{0.97} & \textbf{1.00} & \textbf{1.00} & \textit{0.97} & \textbf{1.00} \\
%\midrule 
EWAF-Overlap-01  & \textbf{1.00} & \textbf{0.99} & \textit{0.98} & \textbf{0.97} & \textbf{1.00} & \textbf{1.00} & \textit{0.97} & \textbf{1.00} \\
%EWAF-Overlap-02  & \textbf{1.00} & \textbf{0.99} & \textit{0.98} & \textbf{0.97} & \textbf{1.00} & \textbf{1.00} & \textit{0.97} & \textbf{1.00} \\
%\midrule 
\midrule 
%WMA-Lev-08-01  & \textbf{1.00} & 0.99 & \textit{0.98} & {0.97} & \textbf{1.00} & \textbf{1.00} & \textit{0.97} & \textbf{1.00} \\
WMA-Lev-08-02  & \textbf{1.00} & 0.99 & \textit{0.98} & {0.97} & \textbf{1.00} & \textbf{1.00} & \textit{0.97} & \textbf{1.00} \\
\bottomrule
\end{tabular}
\end{table}

To summarize, we can see that our approach is successful in two ends. First, learning new patterns leads to improvements compared to using the original patterns, even if not using the implicit feedback of the user to weigh the patterns. Then, by using the implicit feedback to the full extent, we are able to score the patterns to improve the results obtained even further, by ranking the generated questions. However, considering the obtained results, although probable, we cannot claim that questions needing major fixes result from worse patterns, while questions not requiring much editing come from well behaved patterns.

The results obtained improved across all metrics for different cuts of $N$ (5, 10, 20), for the configurations shown. Even with semantic metrics, which are not as discriminative, the strategy showed some improvements. In addition, both ROUGE and METEOR show gains when compared to the two baselines, although at different extents, indicating the effectiveness of this approach. 

BLEU, on the other hand, is the metric with more erratic behavior -- mostly BLEU1 but also applies to BLEU4. It shows less improvements overall for the weighing strategies when compared to both baselines, sometimes even scoring below the baseline. This could be explained by the nature of the BLEU metric, as it uses the whole reference instead of each of the hypotheses in it alone. For example, the question \textit{What designs the palace?}, generated from \textit{The palace was designed by the architects Thomas James Knowles (father and son) and built in 1858, having been commissioned by Sir Francis Cook, Viscount of Monserrate} obtains a high score of $0.75$, when the reference is composed of: \begin{inparaenum}[1)] \item \textit{Who designed the palace?} \item \textit{When was the palace built?} \item \textit{Who commissioned the construction of the palace?} \item \textit{What was Sir Francis Cook noble tittle?}\end{inparaenum}. This means that the baseline is not punished for having ill-formulated questions on the top as much as it should when using BLEU as a metric, contributing to the contradictory results. 

To finalize, Table~\ref{tab:resultsCompared} compiles the results obtained by \sys \texttt{All} and one of the best parameterizations used in the batches (\texttt{EWAF-Overlap-01}). Note that the batch results: \begin{inparaenum} \item only consider the top N questions; \item are an average of all batches.\end{inparaenum}

\begin{table}[t!]
\centering
\caption{Comparison of results obtained with automatic metrics on \McQGE, for H\&S, D\&A, \sys All, and the batched runs.}
\label{tab:resultsCompared}
\small
\begin{tabular}{lrrrrrrrr}
\toprule
& ROUGE & METEOR & BLEU1 & BLEU4 & EACS & GMS & STCS & VECS\\
\midrule
H\&S & \textbf{69.00}  & 46.38  & 83.71 & \textbf{45.56}  & 92.51 & \textbf{86.11} & 73.26 & 77.92 \\
\du & 63.71  & 37.58  & 77.40 & 26.63 &  \textbf{92.52} & 85.51 &  \textbf{74.47} & 77.54\\
\midrule
\sys All & 63.13 & 41.09 & 77.90 & 34.33 & 90.52 & 83.97  & 67.60& {77.90}\\
\midrule
@5 size 7 & 65.92 & 45.20 & 78.95 & 44.16 & 90.01 & 84.10 & 72.92 & 79.82\\
@10 size 7 & 63.79 & 42.04 & 78.83 & 34.67 & 90.23 & 83.76 & 72.00 & 79.21\\
@20 size 7 & 61.74 & 38.91 & 79.32 & 31.05 & 89.96 & 83.92 & 69.74 & 78.28\\
@5 size 10 & 65.93 & 47.97 & 77.04 & 47.18 & 91.28 & 91.28 & 75.04 & 79.56 \\
@10 size 10 & 66.16 & 44.14 & 80.26 & 40.07 & 91.28 & 91.28 & 72.49 & 79.79\\
@20 size 10 & 66.81 & 42.47 & 82.32 & 35.43 & 91.60 & 91.60 & 70.11 & 80.67\\
@5 size 12 & 61.12 & 41.53 & 74.22 & 33.16 & 89.86 & 81.88 & 75.94 & 77.42\\
@10 size 12 & 58.06 & 38.26 & 72.69 & 32.61 & 88.98 & 81.57 & 69.96 & 75.77\\
@20 size 12 & 57.26 & 38.34 & 73.02 & 28.76 & 89.42 & 82.67 & 65.52 & 76.51\\
\bottomrule
\end{tabular}
\end{table}

\section{Evolution over Batches}

As seen before, results show gains in all metrics against the baselines, although to different extents, proving that the technique is effective, even for different batching sizes. However, we must note that there is a clear trade-off between the batches' sizes: if they are too small, there is not enough data to learn new patterns and update the weights; if they are too large, there is more data to learn the weights, but less time to see the impact of that training.

The results presented in the previous section respect to the average results across all but the first batch, but do not show how \sys performs along time. Tables~\ref{tab:evolBatches12} to~\ref{tab:evolBatches7} show the number of patterns and questions per batch for the different runs, along with the number of questions discarded by the expert, and the average normalized Levenshtein distance between the generated questions and their corrections. When looking at Table~\ref{tab:evolBatches7}, concerning 10 batches of size 7, we can see that the sentences in the batches influence the number of questions generated: it appears that some sentences are better sources of questions than others, which will lead to different outcomes, including not generating any questions or less than 20 questions (highlighted), which means top 20 results are meaningless for those batches. 

Nevertheless, results point to a overall increase of patterns and generated questions, with the number of questions discarded increasing at first but then diminishing with time. The edit cost, measured by normalized Levenshtein, seems to be more or less constant across all batches but the first, with tendency to increase over time.

\begin{table}[t!]
\centering
\caption{Number of patterns and questions per batch, along with the number of questions discarded by the expert, and the average question editing, for batches of size 12 (6 batches), on \McQGE.}
\label{tab:evolBatches12}
\begin{tabular}{lrrrrrrr}
\toprule
\textbf{Size 12}  & \textbf{1} & \textbf{2} & \textbf{3} & \textbf{4} & \textbf{5} & \textbf{6} \\
\cmidrule{1-7}
\textbf{Patterns} & 11 & 24 & 33 & 39 & 47 & 42\\
\hfill{}New & - & 16 & 10 & 11 & 17 & 8\\
\textbf{Questions} & 132 & 176 & 266 & 267 & 191 & 503\\
\hfill{}Unique & 52 & 41 & 54 & 72 & 77 & 111\\
\textbf{Discarded} & 23 & 74 & 55 & 71 & 26 & 31\\
\hfill{}\% & 17.42 & 42.05 & 20.68 & 26.59 & 13.61 & 6.16\\
\textbf{Edit Avg} & 0.36 & 0.40 & 0.40 & 0.45 & 0.43 & 0.44\\
\bottomrule
\end{tabular}
\end{table}

\begin{table}[t!]
\centering
\caption{Number of patterns and questions per batch, along with the number of questions discarded by the expert, and the average question editing, for batches of size 10 (7 batches), on \McQGE.}
\label{tab:evolBatches10}
\begin{tabular}{lrrrrrrr}
\toprule
\textbf{Size 10} & \textbf{1} & \textbf{2} & \textbf{3} & \textbf{4} & \textbf{5} & \textbf{6} & \textbf{7}  \\
\cmidrule{1-8}
\textbf{Patterns} & 11 & 24 & 22 & 28 & 30 & 36 & 33\\
\hfill{}New & - & 16 & 6 & 9 & 6 & 13 & 7\\
\textbf{Questions} & 121 & 175 & 67 & 215 & 155 & 108 & 379\\
\hfill{}Unique & 50 & 35 & 25 & 54 & 56 & 52 & 95\\
\textbf{Discarded} & 23 & 74 & 1 & 56 & 27 & 11 & 28\\
\hfill{}\% & 19.01 & 42.29 & 1.49 & 26.05 & 17.42 & 10.19 & 7.39\\
\textbf{Edit Avg} & 0.35 & 0.45 & 0.42 & 0.37 & 0.43 & 0.45 & 0.44\\
\bottomrule
\end{tabular}
\end{table}

\begin{table}[t!]
\centering
\caption{Number of patterns and questions per batch, along with the number of questions discarded by the expert, and the average question editing, for batches of size 7 (10 batches), on \McQGE.}
\label{tab:evolBatches7}
\begin{tabular}{lrrrrrrrrrr}
\toprule
\textbf{Size 7}  & \textbf{1} & \textbf{2} & \textbf{3} & \textbf{4} & \textbf{5} & \textbf{6} & \textbf{7} & \textbf{8} & \textbf{9} & \textbf{10}\\
\midrule
\textbf{Patterns} & 11 & 26 & 21 & 20 & 28 & 31 & 30 & 32 & 32 & 43\\
\hfill{}New & - & 26 & 3 & 1 & 8 & 6 & 2 & 6 & - & 16\\
\textbf{Questions} & 210 & 237 & 37 & 46 & 213 & 85 & 120 & 0 & 236 & 369\\
\hfill{}Unique & 50 & 42 & \textbf{10} & \textbf{14} & 51 & 24 & 37 & - & 70 & 90\\
\textbf{Discarded} & 36 & 108 & 3 & 1 & 62 & 24 & 34 & - & 7 & 27\\
\hfill{}\% & 17.14 & 45.57 & 8.11 & 2.17 & 29.11 & 28.24 & 28.33 & - & 2.97 & 7.32\\
\textbf{Edit Avg} & 0.36 & 0.40 & 0.36 & 0.37 & 0.36 & 0.44 & 0.43 & - & 0.48 & 0.40\\
\bottomrule
\end{tabular}
\end{table}

Given that each batch seems to lead to different performances based on its content, it is also interesting to analyze their results individually, instead of looking at the overall average score. As noted before, with semantic metrics it is hard to understand how much the performance changes, so we look at the lexical metrics only (ROUGE, METEOR, BLEU1, BLEU4). 

\begin{figure*}[t!]
  \centering
  \begin{subfigure}[b]{\textwidth}
    \includegraphics[width=\linewidth]{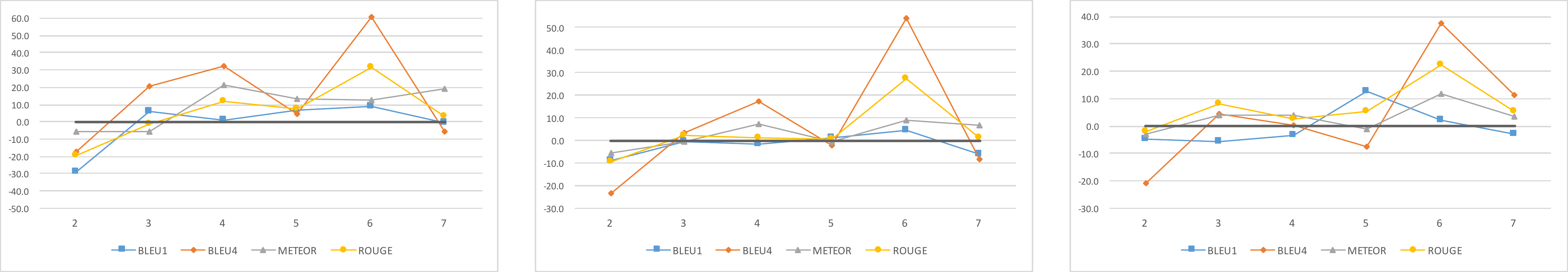}
   \end{subfigure}
\begin{subfigure}[b]{\textwidth}
  \centering
    \includegraphics[width=\linewidth]{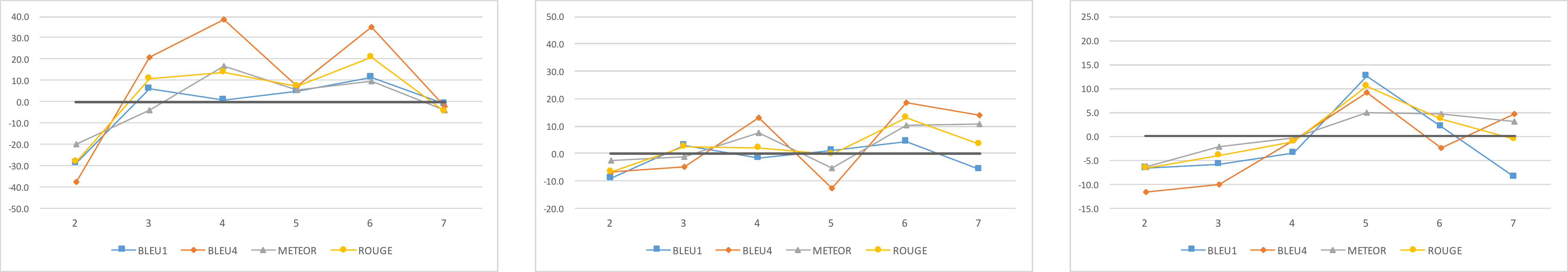}
   \end{subfigure}
     \caption{Difference between the \texttt{baseline} score and the score obtained by \texttt{EWAF-Overlap-01} (top) and \texttt{WMA-Lev-08-02} (bottom), over all but the first batch. Analysis for batches of size 10 (7 batches), at top 5 (left), 10 (middle), and 20 (right).}
    \label{fig:evol10}
\end{figure*}

\begin{figure*}[t!]
  \centering
\begin{subfigure}[b]{\textwidth}
    \includegraphics[width=\linewidth]{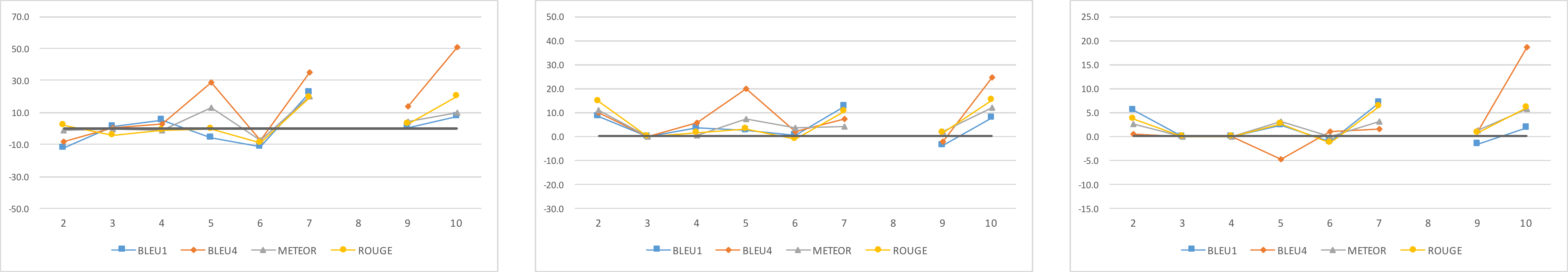}
   \end{subfigure}
\begin{subfigure}[b]{\textwidth}
  \centering
    \includegraphics[width=\linewidth]{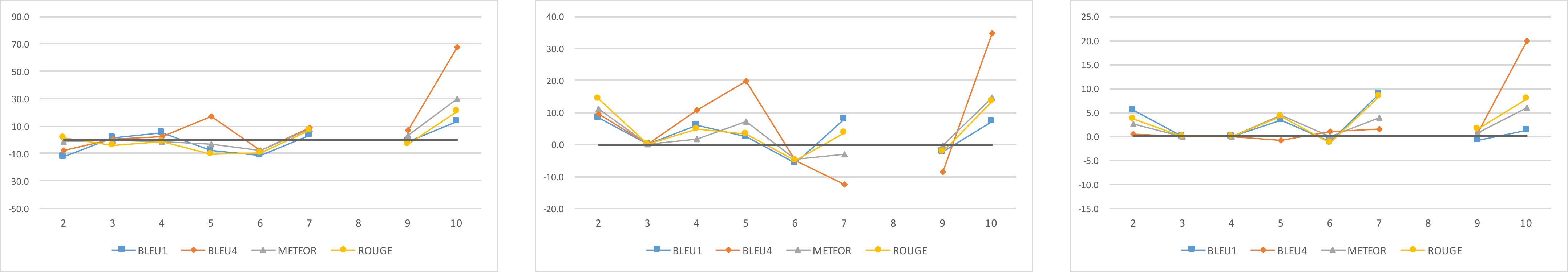}
   \end{subfigure}
     \caption{Difference between the \texttt{baseline} score and the score obtained by \texttt{EWAF-Overlap-01} (top) and \texttt{WMA-Lev-08-02} (bottom), over all but the first batch. Analysis for batches of size 7 (10 batches), at top 5 (left), 10 (middle), and 20 (right).}
    \label{fig:evol7}

\end{figure*}

\begin{figure*}[t!]
  \centering
\begin{subfigure}[b]{\textwidth}
    \includegraphics[width=\linewidth]{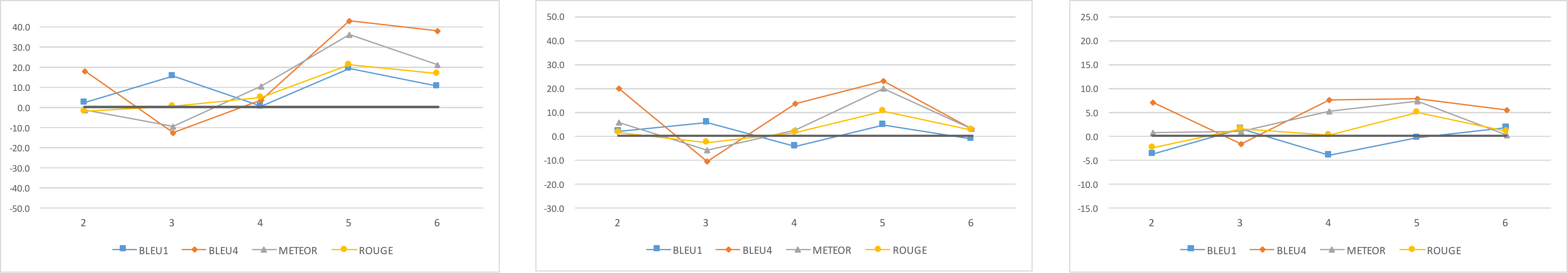}
   \end{subfigure}
\begin{subfigure}[b]{\textwidth}
  \centering
    \includegraphics[width=\linewidth]{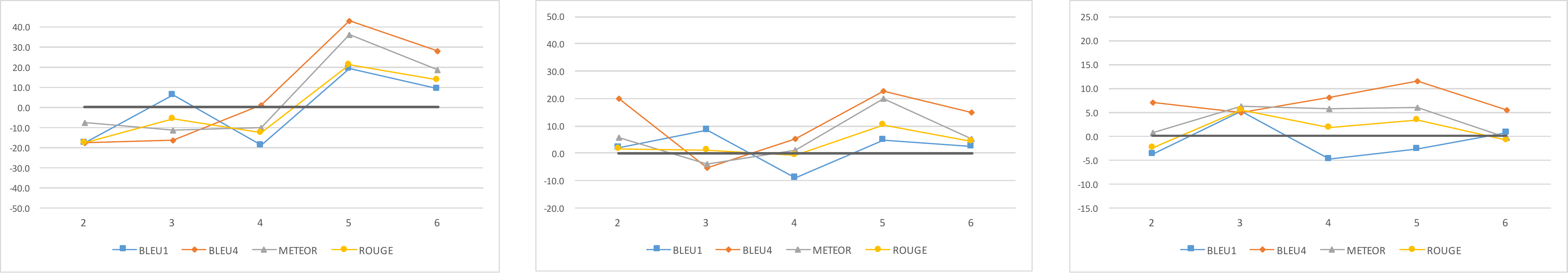}
   \end{subfigure}
     \caption{Difference between the \texttt{baseline} score and the score obtained by \texttt{EWAF-Overlap-01} (top) and \texttt{WMA-Lev-08-02} (bottom), over all but the first batch. Analysis for batches of size 12 (6 batches), at top 5 (left), 10 (middle), and 20 (right).}
    \label{fig:evol12}
\end{figure*}

In Figures~\ref{fig:evol10} to~\ref{fig:evol12} it is shown the evolution for \texttt{EWAF-Overlap-01} and \texttt{WMA-} \texttt{Lev-08-02} for all 3 runs -- batches of size 10, 7, and 12  (7, 10 and 6 batches, respectively). The figures presented depict the difference in score between the \texttt{baseline} scores and the ones obtained with the given strategy, for each of the batches individually, that is, the graph is not a cumulative evolution of the scores. The figures are sequentially in pairs; for instance, for batches of size 10, Figure~\ref{fig:evol10} shows the evolution for \texttt{EWAF-Overlap-01} and \texttt{WMA-Lev-08-02} at top and bottom, respectively, for top $N$ 5, 10, and 20, from left to right.

In all cases we can see batches that surpass the \texttt{baseline} and others that do not. However, we can see that the gains surpass the losses, seen by how often and by how much the lines stand over the zero axis.

We take a closer look at Figure~\ref{fig:evol7}, which depicts the evolution for batches of size 7 (10 batches). We can see, at top 20, that almost no loss is happening (almost all points are positive, i.e., they are above the axis). In one hand, because the batches generate less questions (sometimes close or even less than 20), there is less room to lose points to the \texttt{baseline}. On the other hand, that room still exists and it is being avoided, that is, the system is able to consistently improve the ordering of the questions without committing mistakes. However, this does not happen for smaller values of $N$. For instance, top 5 shows more losses except for a few batches. 

We believe larger batches have greater impact on the performance of these strategies. The sweet spot might not be easy to find but, if the corpus is large (\McQGE is large on the reference side, but not in number of sentences), batches of 10 sentences should be good enough to make a difference. Additionally, the contents of the batches can impact the perceived performance of these strategies but, in the end, the gains outweigh the losses, as seen by the overall scores. Therefore, one cannot make conclusions based on a single iteration but should, rather, look to apply these strategies on the long term. In sum, we found out that the batching experiment had a few more variables that make it hard to evaluate our hypothesis, but results still point towards the scores of the patterns being positively impactful, which means our hypothesis holds: patterns with better scores are, by definition, better behaved, and those are used to rank better questions to the top.

%%%%%%%%%%%%%%%%%%%%%%%%%%%%%%%%%%%%%
%%%%%%%%%%%%%%%%%%%%%%%%%%%%%%%%%%%%%

\section{Ordering of Input}

The experiments presented split the corpus in the same order, which may introduce bias into the equation. It is feasible that a lucky ordering of inputs lead to learning better weights, showing an apparent improvement. In this section we repeat the experiment for batches created from a shuffled version of the corpus. We chose 7 new batches of 10 sentences each for this second version of the experiment.

\begin{table}[t!]
\centering
\caption{Number of patterns and questions per batch, along with the number of questions discarded by the expert, and the average question editing, for batches of size 10 (7 batches), on \McQGE (shuffled order).}
\label{tab:evolBatches10shuf}
\begin{tabular}{lrrrrrrr}
\toprule
\textbf{Size 10} (v2)  & \textbf{1} & \textbf{2} & \textbf{3} & \textbf{4} & \textbf{5} & \textbf{6} & \textbf{7}\\
\cmidrule{1-8}
\textbf{Patterns} & 11 & 23 & 31 & 35 & 38 & 53 & 41\\
\hfill{}New & - & 13 & 12 & 9 & 9 & 20 & 7\\
\textbf{Questions} & 51 & 130 & 268 & 290 & 432 & 123 & 519\\
\hfill{}Unique & 32 & 52 & 98 & 87 & 111 & 26 & 115\\
\textbf{Discarded} & 4 & 25 & 74 & 25 & 50 & 41 & 84\\
\hfill{}\% & 7.84 & 19.23 & 27.61 & 8.62 & 11.57 & 33.33 & 16.18\\
\textbf{Edit Avg} & 0.40 & 0.42 & 0.46 & 0.37 & 0.44 & 0.49 & 0.42\\
\bottomrule
\end{tabular}
\end{table}

\begin{figure}[t!]
  \centering
  \begin{subfigure}[b]{\textwidth}
    \includegraphics[width=\linewidth]{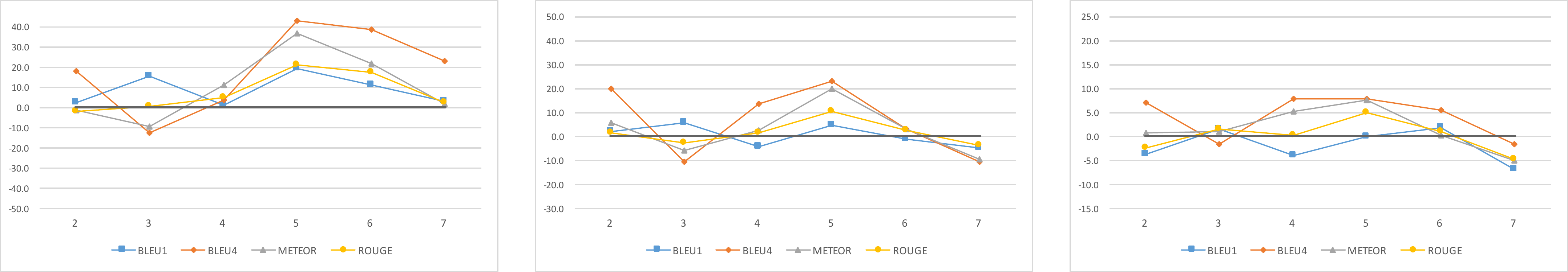}
   \end{subfigure}
\begin{subfigure}[b]{\textwidth}
  \centering
    \includegraphics[width=\linewidth]{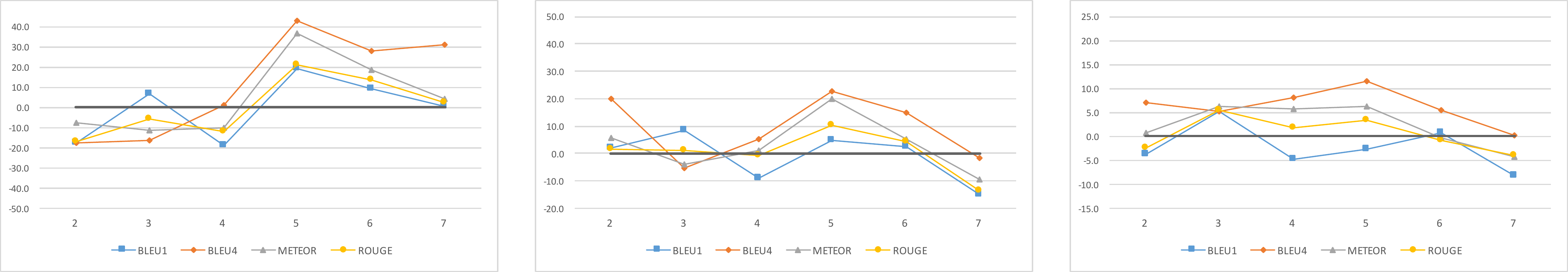}
   \end{subfigure}
     \caption{Difference between the baseline score and the score obtained by \texttt{EWAF-Overlap-01} (top) and \texttt{WMA-Lev-08-02} (bottom), over all batches but the first. Analysis for 7 batches (shuffled), at top 5 (left), 10 (middle), and 20 (right).}
    \label{fig:evol10shuf}
\end{figure}

Results are presented in Table~\ref{tab:resultsIter10-5shuf}. Although with different gains, we can see that there are gains for all metrics for the same strategies and configurations, for all values of $N$ (with less effectiveness on top 20).

The evolution of the learning process, however, is completely different, as seen in Table~\ref{tab:evolBatches10shuf} and Figure~\ref{fig:evol10shuf}. Comparing both runs of 7 batches we can see that the trend is completely different, which means that the performance of the system also depends on the content seen. As suggested before, the ordering impacts the content of the batches and, directly, their outcomes. Thus, these differences are expected. However, in the end, the average results improve when compared to the baselines. Therefore, it is hard to reach a conclusion on how the system evolves over time, and if later batches perform better, as they depend on the previous ones, but we can at least say that the overall performance is improved with these strategies, independently of the ordering.

\begin{table}[t!]
\centering
\caption{Comparison of the weighing strategies against the baselines, measured by automatic metrics on \McQGE (shuffled order), at top 5, 10, and 20. Scores normalized by the best score obtained in each metric. Overall results for batches of size 10 (7 batches) -- averaged on all but first batch.}
\label{tab:resultsIter10-5shuf}
\small
\begin{tabular}{lrrrrrrrr}
\toprule
@5 & ROUGE & METEOR & BLEU1 & BLEU4 & EACS & GMS & STCS & VECS\\
\midrule
\midrule
Original Patterns  & 0.91 & 0.87 & 0.91 & 0.61 & \textbf{1.00} & 0.98 & 0.89 & 0.95\\
\midrule 
Baseline  & \textit{0.89} & \textit{0.79} & \textit{0.90} & 0.65 & \textit{0.99} & 0.98 & 0.92 & \textit{0.94}\\
\midrule 
\midrule 
%EWAF-Lev-01  & 0.98 & 0.99 & 0.98 & 0.99 & \textbf{1.00} & 0.99 & \textbf{0.98} & 0.99\\
%EWAF-Lev-02  & 0.98 & 0.99 & 0.98 & 0.99 & \textbf{1.00} & 0.99 & \textbf{0.98} & 0.99\\
%\midrule
EWAF-Overlap-01  & \textbf{1.00} & \textbf{1.00} & \textbf{1.00} & \textbf{1.00} & \textbf{1.00} & \textbf{1.00} & 0.97 & \textbf{1.00}\\
%EWAF-Overlap-02  & 0.98 & 0.99 & 0.98 & 0.99 & \textbf{1.00} & 0.99 & \textbf{0.98} & 0.99\\
%\midrule  
\midrule 
%WMA-Lev-08-01  & \textit{0.87} & 0.91 & \textit{0.86} & 0.76 & \textit{0.98} & \textit{0.96} & 0.96 & 0.96\\
WMA-Lev-08-02  & \textit{0.89} & 0.90 & \textit{0.90} & 0.86 & \textit{0.99} & \textit{0.97} & 0.95 & 0.97\\
%WMA-Lev-09-01  & \textbf{0.99} & \textbf{0.94} & \textbf{0.95} & 0.85 & \textbf{1.00} & \textbf{1.00} & \textbf{1.00} & \textbf{1.00}\\
%WMA-Lev-09-02  & 0.92 & 0.91 & \textit{0.88} & \textbf{0.97} & \textit{0.98} & 0.98 & 0.97 & 0.97\\
%\midrule 
%\midrule  
%WMA-Overlap-08-01-B03  & \textit{0.85} & \textit{0.78} & \textit{0.87} & \textit{0.39} & \textit{0.98} & \textit{0.96} & \textbf{0.93} & \textit{0.94}\\
\midrule
\midrule
@10 & ROUGE & METEOR & BLEU1 & BLEU4 & EACS & GMS & STCS & VECS\\
\midrule
Original Patterns  & 0.96 & 0.94 & 0.99 & 0.79 & 0.99 & 0.98 & 0.89 & 0.96\\
\midrule 
%\midrule
Baseline  &\textit{0.94} & \textit{0.90} & \textit{0.97} & \textit{0.77} & 0.99 & 0.99 & 0.92 & 0.96\\
\midrule 
\midrule 
%EWAF-Lev-01  & \textbf{1.00} & \textbf{1.00} & \textbf{0.99} & \textbf{1.00} & \textbf{1.00} & \textbf{1.00} & \textbf{0.99} & \textbf{1.00}\\
%EWAF-Lev-02  & \textbf{1.00} & \textbf{1.00} & \textbf{0.99} & \textbf{1.00} & \textbf{1.00} & \textbf{1.00} & \textbf{0.99} & \textbf{1.00}\\
%\midrule 
EWAF-Overlap-01  & 0.97 & 0.96 & \textit{0.97} & 0.91 & 0.99 & 0.99 & 0.97 & 0.99\\
%EWAF-Overlap-02  & 0.99 & 0.99 & \textit{0.98} & 0.97 & 0.99 & 0.99 & 0.98 & \textbf{1.00}\\
%\midrule 
\midrule 
%WMA-Lev-08-01  & \textit{0.95} & 0.96 & \textit{0.95} & 0.95 & {0.99} & 0.98 & 0.99 & 0.99\\
WMA-Lev-08-02  & \textit{0.95} & \textbf{0.97} & \textit{0.95} & \textbf{0.98} & {0.99} & 0.99 & 0.99 & 0.99\\
\midrule
\midrule
@20 & ROUGE & METEOR & BLEU1 & BLEU4 & EACS & GMS & STCS & VECS\\
\midrule
Original Patterns  & \textbf{1.00} & 0.97 & \textbf{1.00} & 0.89 & 0.99 & 0.98 & 0.93 & 0.97\\
\midrule 
%\midrule
Baseline  & \textit{0.97} & \textit{0.94} & \textit{0.96} & \textit{0.83} & 0.99 & 0.99 & 0.96 & 0.98\\
\midrule 
\midrule 
%EWAF-Lev-01  & \textit{0.97} & \textbf{1.00} & \textit{0.94} & \textbf{0.96} & 0.99 & \textbf{1.00} & \textbf{0.99} & \textbf{0.99}\\
%EWAF-Lev-02  & \textit{0.97} & \textbf{1.00} & \textit{0.94} & \textbf{0.96} & 0.99 & \textbf{1.00} & \textbf{0.99} & \textbf{0.99}\\
%\midrule 
EWAF-Overlap-01  & \textit{0.97} & 0.98 & \textit{0.94} & 0.94 & 0.99 & 0.99 & \textbf{0.99} & \textbf{0.99}\\
%EWAF-Overlap-02  & \textit{0.97} & 0.98 & \textit{0.94} & 0.94 & 0.99 & 0.99 & \textbf{0.99} & \textbf{0.99}\\
%\midrule 
\midrule 
%WMA-Lev-08-01  & \textit{0.97} & 0.98 & \textit{0.93} & 0.99 & \textbf{1.00} & \textbf{1.00} & \textbf{1.00} & 0.99\\
WMA-Lev-08-02  & \textit{0.98} & \textbf{0.99} & \textit{0.93} & \textbf{1.00} & \textbf{1.00} & \textbf{1.00} & \textbf{1.00} & \textbf{1.00}\\
\bottomrule
\end{tabular}
\end{table}

%%%%%%%%%%%%%%%%%%%%%%%%%%%%%%
%%%%%%%%%%%%%%%%%%%%%%%%%%%%%%
%%%%%%%%%%%%%%%%%%%%%%%%%%%%%%
%%%%%%%%%%%%%%%%%%%%%%%%%%%%%%
%%%%%%%%%%%%%%%%%%%%%%%%%%%%%%

\section{Learning New Seeds on SQuAD}
\label{sec:weightSquad}
SQuAD \cite{Rajpurkar16} is a widely used corpus, mainly in \ac{QA} and \ac{QG} settings. SQuAD is a large corpus, but as a reference it has a few limitations (only a single reference hypothesis associated to each sentence, for instance), reason why we performed the whole experiment in this work using \McQGE, as it becomes easier to use automatic metrics to perform the evaluation. However, to understand how \sys would perform in a different corpus, we decided to repeat the experiment on SQuAD. 

Due to the high cost of correcting questions, this time we executed a single attempt using bathes of size 10, showing the results to the two best weighing techniques also resported in the previous sections (\texttt{EWAF-Overlap-01} and \texttt{WMA-Lev-08-02}). The whole process is identical to the one presented in Section~\ref{sec:es}. Because SQuAD is a larger corpus, it will lead to many batches of size 10. On \McQGE it is possible to do 7 batches, so here we extended to 10 batches in order to explore what happens when more batches are possible. Note that using automatic metrics with SQuAD is suboptimal, because having only one question hypothesis in the reference punishes not generating a similar question to the one expected.

\begin{figure*}[t!]
  \centering
  \begin{subfigure}[b]{\textwidth}
    \includegraphics[width=\linewidth]{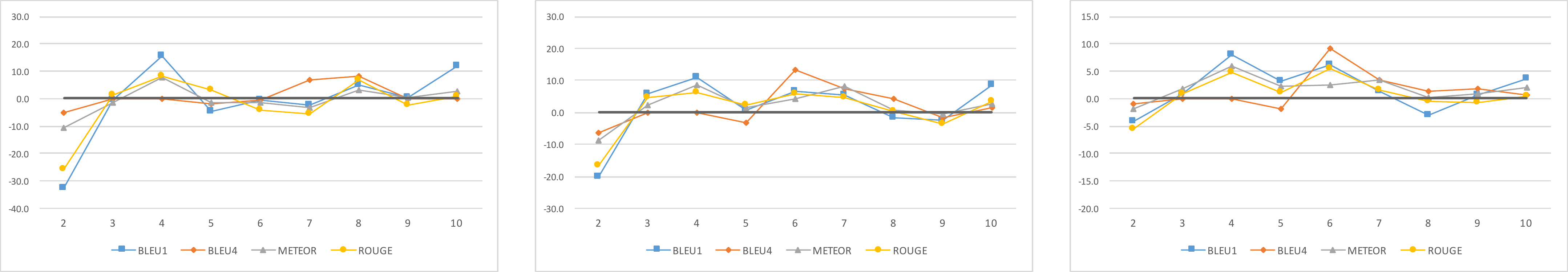}
   \end{subfigure}
\begin{subfigure}[b]{\textwidth}
  \centering
    \includegraphics[width=\linewidth]{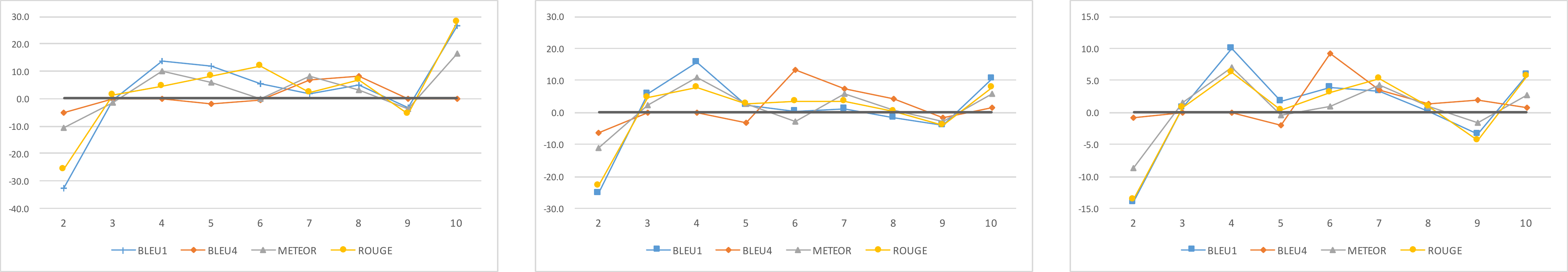}
   \end{subfigure}
     \caption{Difference between the baseline score and the score obtained by \texttt{EWAF-Overlap-01} (top) and \texttt{WMA-Lev-08-02} (bottom), over all batches but the first. Analysis for 10 batches of size 10 on SQuAD, at top 5 (left), 10 (middle), and 20 (right).}
    \label{fig:evolSQuAD}
\end{figure*}

Table~\ref{tab:resultsIterSquad-5} shows the results obtained for both weighing strategies (\texttt{EWAF-Overlap-01} and \texttt{WMA-Lev-08-02}), using the same thresholds for $N$ -- top 5, 10, and 20. Table~\ref{tab:evolBatchesSQuAD} contains the statistics on the evolution of the batches regarding the number of patterns, questions generated and discarded, and the average edit cost measured by normalized Levenshtein. Figure~\ref{fig:evolSQuAD} depicts the gains obtained by each of the strategies against the \texttt{baseline}, for each batch but the first.

\begin{table}[t!]
\centering
\caption{Comparison of the weighing strategies against the baseline, measured by automatic metrics on SQuAD, at top 5, 10, and 20. Scores normalized by the best score obtained in each metric. Overall results for batches of size 10 (10 batches) -- averaged on all but first batch.}
\label{tab:resultsIterSquad-5}
\small
\begin{tabular}{lrrrrrrrr}
\toprule
@5 & ROUGE & METEOR & BLEU1 & BLEU4 & EACS & GMS & STCS & VECS\\
\midrule
\midrule
%Original Patterns  &  \\
%\midrule 
%\midrule 
Baseline  & 0.87 & 0.81 & 0.86 & 0.26 & 0.99 & 0.97 & 0.88 & \textbf{1.00} \\
\midrule 
\midrule 
EWAF-Overlap-01 & \textit{0.80} & \textit{0.79} & \textit{0.82} & 0.41 & \textit{0.98} & \textit{0.95} & \textit{0.87} & \textbf{1.00}\\
WMA-Lev-08-02 & \textbf{1.00} & \textbf{1.00} & \textbf{1.00} & \textbf{1.00} & \textbf{1.00} & \textbf{1.00} & \textbf{1.00} & 0.97\\
\midrule
\midrule
@10 & ROUGE & METEOR & BLEU1 & BLEU4 & EACS & GMS & STCS & VECS\\
\midrule
%Original Patterns  &  \\
%\midrule 
%\midrule 
Baseline  & 0.97 & 0.87 & 0.93 & 0.66 & 0.98 & \textbf{1.00} & 0.91 & 0.94 \\
\midrule 
\midrule 
EWAF-Overlap-01 & \textbf{1.00} & \textbf{1.00} & \textbf{1.00} & \textbf{1.00} & \textbf{1.00} & \textbf{1.00} & 0.99 & \textbf{1.00}\\
WMA-Lev-08-02 & 0.98 & 0.95 & 0.95 & 0.89 & 0.99 & \textbf{1.00} & \textbf{1.00} & \textbf{1.00}\\
\midrule
\midrule
@20 & ROUGE & METEOR & BLEU1 & BLEU4 & EACS & GMS & STCS & VECS\\
\midrule
%Original Patterns  &  \\
%\midrule 
%\midrule 
Baseline  & 0.97 & 0.87 & 0.91 & 0.67 & 0.99 & \textbf{1.00} & 0.94 & 0.96 \\
\midrule 
\midrule 
EWAF-Overlap-01 & \textbf{1.00} & \textbf{1.00} & \textbf{1.00} & \textbf{1.00} & \textbf{1.00} & \textbf{1.00} & \textbf{1.00} & \textbf{1.00}\\
WMA-Lev-08-02 & 0.99 & 0.92 & 0.95 & 0.66 & \textbf{1.00} & \textbf{1.00} & \textbf{1.00} & 0.99\\
\bottomrule
\end{tabular}
\end{table}

\begin{table}[t!]
\centering
\caption{Number of patterns and questions per batch, along with the number of questions discarded by the expert, and the average question editing, for batches of size 10 (10 batches), on SQuAD.}
\label{tab:evolBatchesSQuAD}
\begin{tabular}{lrrrrrrrrrr}
\toprule
\textbf{Size 7}  & \textbf{1} & \textbf{2} & \textbf{3} & \textbf{4} & \textbf{5} & \textbf{6} & \textbf{7} & \textbf{8} & \textbf{9} & \textbf{10}\\
\midrule
\textbf{Patterns} & 11 & 21 & 38 & 44 & 66 & 81 & 108 & 116 & 121 & 126 \\
\textbf{Questions} & 41 & 142 & 125 & 267 & 394 & 435 & 541 & 265 & 457 & 411 \\
\hfill{}Unique & 25 & 82 & 40 & 108 & 107 & 103 & 144 & 49 & 95 & 92\\
\textbf{Discarded} & 6 & 27 & 10 & 33 & 63 & 40 & 204 & 45 & 51 & 101 \\
\hfill{}\% & 14.63 & 19.01 & 8.00 & 12.36 & 15.99 & 9.20 & 37.71 & 16.98 & 11.16 & 24.57 \\
\textbf{Edit Avg.} & 0.43 & 0.49 & 0.40 & 0.39 & 0.39 & 0.37 & 0.36 & 0.47 & 0.42 & 0.49\\
\bottomrule
\end{tabular}
\end{table}

Results show a similar trend to the previous experiment (with exception for \texttt{EWAF-\\Overlap-01} at top 5), both in overall results and per batch, with gains in most batches that overcome those with losses, corroborating the finds in last sections.

%%%%%%%%%%%%%%%%%%%%%%%%%%%%%%
%%%%%%%%%%%%%%%%%%%%%%%%%%%%%%
%%%%%%%%%%%%%%%%%%%%%%%%%%%%%%
%%%%%%%%%%%%%%%%%%%%%%%%%%%%%%
%%%%%%%%%%%%%%%%%%%%%%%%%%%%%%

\section{Conclusions and Future Work}\label{sec:conc}
In this work we studied the concept of using implicit feedback as means to improve a \ac{QG} system. To accomplish that, we proposed \sys, a pattern-based \ac{QG} system that automatically learns patterns from seeds composed of Q/A pairs and a sentence where the answer can be found. \sys uses a flexible pattern matching strategy based on lexical, syntactic, and semantic cues to create new questions from unseen sentences. Being able to automatically learn patterns from seeds is the reason why it is possible to incorporate feedback that comes directly from the user into the system. Users typically have to correct the generated questions to be used in their application, and this effort is usually wasted. These corrections can be useful, and are used in this work in two ways: first, they are used as trustful sources of new seeds to learn new patterns from; secondly, the correction of the questions themselves are used as a function of how well the generation process performed. This, indirectly, establishes how good the pattern that generated a given question is, and that is used to score the pattern and, therefore, future questions coming from that same pattern. 

We studied how \sys evolved over time, by batching \McQGE corpus and evaluating the performance of the system as new patterns were learned and past ones were weighed. Results with automatic metrics showed that \sys was able to obtain higher overall scores across all metrics, in different parameterizations for the weighing strategies applied, when compared with the two baselines set: original patterns with no learning, and learning new patterns but not scoring them. The success of this approach was also visible when using SQuAD, although to a different extent due to the nature of that corpus.

We also established that, although overall scores were higher, results for single batches may vary. We showed that sequencing can influence the results obtained on a given batch, and, for that reason, we cannot claim that there is a unique best way to apply these weighing strategies. Rather, results point to overall gains in the long term, independently of the size of the batches and corpora used, with improvements going up from 10\%, depending on the metric and strategy used. Moreover, results suggest that batches of at least size 10 are large enough to use these strategies successfully. 

As for future work, we would like to extend the impact of the user feedback, by having more fine grained scores and how those are incorporated. For instance, the type of matching used could result in different rewards/penalties, and those could also change throughout time instead of being predefined from start. 

Human evaluation could also be conducted to better support our claims. However, we should note that the focus of this work was to evaluate the gains obtained by using the implicit feedback and how that translates on the performance of the system against itself. In addition, \McQGE gives more reliability to using automatic metrics in the evaluation procedure, and, thus, more confidence on the results obtained.

Finally, a similar study conducted on another domain could provide a better analysis on the batching schema to be employed.

\section*{Aknowledgements}
Hugo Rodrigues was supported by the Carnegie Mellon-Portugal program (SFRH/BD/51916/\\2012). This work was also supported by national funds through Funda\c c\~ao para a Ci\^encia e Tecnologia (FCT) with reference UIDB/50021/2020, and by FEDER, Programa Operacional Regional de Lisboa, Agência Nacional de Inovação (ANI), and CMU Portugal, under the project Ref. 045909 (MAIA).

\bibliographystyle{plainnat}
\bibliography{references}

\label{lastpage}

\end{document}